\documentclass{egpubl}
\usepackage{eg2026}
 
\PpConferencePaper        % uncomment for (final) Conference Paper
\CGFStandardLicense
\usepackage[T1]{fontenc}
\usepackage{dfadobe}
\usepackage{amsmath,amssymb,amsfonts}

\usepackage{cite}  % comment out for biblatex with backend=biber
\BibtexOrBiblatex
\electronicVersion
\PrintedOrElectronic
\ifpdf \usepackage[pdftex]{graphicx} \pdfcompresslevel=9
\else \usepackage[dvips]{graphicx} \fi

\usepackage{egweblnk} 
\usepackage[table]{xcolor}
\usepackage{booktabs}
\usepackage{pdfpages}
\definecolor{nice}{RGB}{255,255,153}
\definecolor{good}{RGB}{255,204,153}
\definecolor{best}{RGB}{255,153,153}

\definecolor{darkgreen}{rgb}{0,0,0}
\newcommand{\changed}[1]{\textcolor{darkgreen}{#1}}

\title[GS-2M]%
      {GS-2M: \changed{Material-aware} Gaussian Splatting for \\\changed{High-fidelity} Mesh Reconstruction}

\author[D.\,M. Nguyen et al.]
{
\parbox{\textwidth}
{\centering
D.\,M. Nguyen$^{1}$\orcid{0009-0000-7102-7495},
M. Avenhaus$^{2}$\orcid{0009-0003-5310-8971},
and T. Lindemeier$^{2}$\orcid{0009-0003-7715-8439} 
}
        \\
{\parbox{\textwidth}
{\centering
$^1$Norwegian University of Science and Technology, Norway\\
$^2$Carl Zeiss AG, Germany
}
}
}
\begin{document}

\teaser{
 \includegraphics[width=0.9\linewidth]{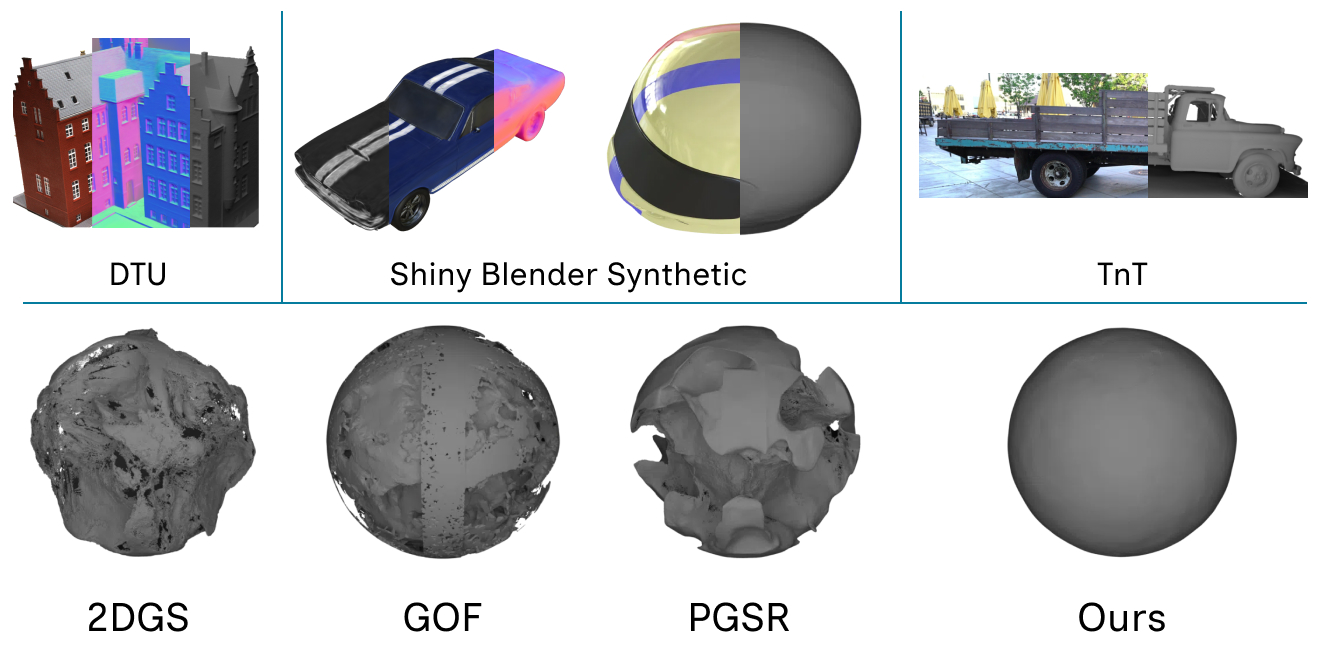}
 \centering
  \caption{Our approach \changed{reinforces} mesh reconstruction \changed{by incorporating} material decomposition into a \changed{joint} optimization framework, \changed{producing} high-fidelity triangle meshes even for reflective surfaces. We validate the effectiveness of our method (top row) with the DTU benchmark (left), the Shiny Blender Synthetic dataset (middle), and the TanksAndTemples dataset (right). We also provide qualitative comparisons (bottom row) between our method and state-of-the-art surface reconstruction methods to highlight the challenge of recovering triangle meshes for reflective objects.}
\label{fig-teaser}
}

\maketitle
%-------------------------------------------------------------------------

\begin{abstract}
   We propose a \changed{material-aware optimization framework} for \changed{high-fidelity} mesh reconstruction from multi-view images based on 3D Gaussian Splatting, referred to as GS-2M. Previous works handle these tasks separately and struggle to reconstruct highly reflective surfaces, often relying on priors from external models to enhance the decomposition results. Conversely, our method addresses these two problems by jointly optimizing attributes relevant to the quality of rendered depth and normals, maintaining geometric details while being resilient to reflective surfaces. Although contemporary works effectively solve these tasks together, they often employ sophisticated neural components to learn scene properties, which hinders their performance at scale. To further eliminate these neural components, we propose a novel roughness supervision strategy based on multi-view photometric variation. When combined with a carefully designed loss and optimization process, our unified framework produces reconstruction results comparable to state-of-the-art methods, delivering \changed{accurate triangle meshes even for reflective surfaces}. We validate the effectiveness of our approach with widely used datasets from previous works and qualitative comparisons with state-of-the-art surface reconstruction methods. Project page: \URL{https://ndming.github.io/publications/gs2m/}.
%-------------------------------------------------------------------------
%  ACM CCS 1998
%  (see https://www.acm.org/publications/computing-classification-system/1998)
% \begin{classification} % according to https://www.acm.org/publications/computing-classification-system/1998
% \CCScat{Computer Graphics}{I.3.3}{Picture/Image Generation}{Line and curve generation}
% \end{classification}
%-------------------------------------------------------------------------
%  ACM CCS 2012
%   (see https://www.acm.org/publications/class-2012)
%The tool at \url{http://dl.acm.org/ccs.cfm} can be used to generate
% CCS codes.
%Example:
% \begin{CCSXML}
% <ccs2012>
% <concept>
% <concept_id>10010147.10010371.10010352.10010381</concept_id>
% <concept_desc>Computing methodologies~Collision detection</concept_desc>
% <concept_significance>300</concept_significance>
% </concept>
% <concept>
% <concept_id>10010583.10010588.10010559</concept_id>
% <concept_desc>Hardware~Sensors and actuators</concept_desc>
% <concept_significance>300</concept_significance>
% </concept>
% <concept>
% <concept_id>10010583.10010584.10010587</concept_id>
% <concept_desc>Hardware~PCB design and layout</concept_desc>
% <concept_significance>100</concept_significance>
% </concept>
% </ccs2012>
% \end{CCSXML}


\end{abstract}

\hypersetup{colorlinks=true,
  linkcolor=blue,
  urlcolor=blue,
  citecolor=blue}
%-------------------------------------------------------------------------

\section{Introduction}
Reconstructing triangle meshes from multi-view images is a highly relevant problem within the visual computing domain, allowing the acquisition of 3D models from photos without the need for laborious manual work. With the advent of Neural Radiance Fields (NeRF) \cite{mildenhall2020nerf}, numerous neural implicit surface reconstruction methods have been introduced \cite{yariv2021volume, wang2021neus, Zhang2022regsdf, darmon2022neuralwarp, wang2023neus2, li2023neuralangelo}; however, they often require hours of training on high-end GPUs to achieve sufficient output quality, hindering their applicability in practice. While methods exist to reduce training time \cite{mueller2022instant, wang2023neus2}, they are limited and often come at the expense of reconstruction performance.

% \cite{guedon2023sugar, Dai2024GaussianSurfels, Huang2DGS2024, Yu2024GOF, wang2024GausSurf, Chen2025PGSR, guedon2025milo}
% \cite{Jiang2024GaussianShader, Ye2024Deferred3DGS, Bi2024gs3, ye2024geosplatting, Liang2024gsir, Wu2025DeferredGS, Shi2025GIR, zhou2025rtrgs, zhu2025gsror2, glossygs, Zhang2025refgs}
%  \cite{Huang2DGS2024, Yu2024GOF, wang2024GausSurf, Chen2025PGSR}
% \cite{Jiang2024GaussianShader, Ye2024Deferred3DGS, Bi2024gs3, ye2024geosplatting, Liang2024gsir, Wu2025DeferredGS, Shi2025GIR, zhou2025rtrgs}
% \cite{jensen2014dtu, Knapitsch2017tnt}
% realizing the significance of the underlying surface geometry and

\begin{figure}
    \centering
    \includegraphics[width=1\linewidth]{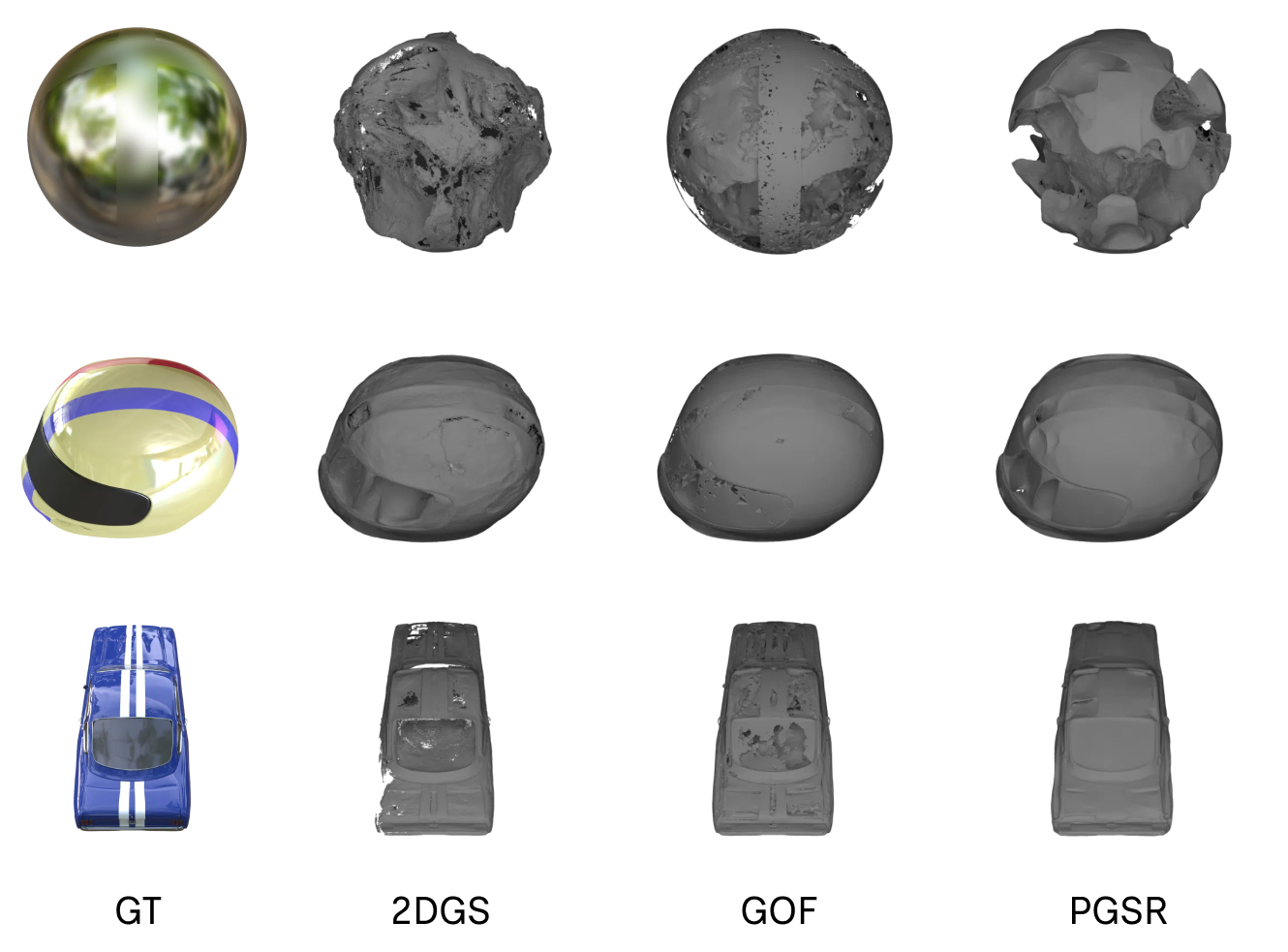}
    \caption{The reconstructed meshes of reflective objects taken from the Shiny Blender Synthetic dataset \cite{Verbin2022refnerf}, experimented on recent SoTA surface reconstruction methods: 2DGS \cite{Huang2DGS2024}, GOF \cite{Yu2024GOF}, and PGSR \cite{Chen2025PGSR}. Due to the lack of appearance modeling, these methods often sacrifice geometric details for view-dependent effects caused by highly specular materials, resulting in non-watertight or distorted meshes.}
    \label{fig-sota-reflective}
\end{figure}

Most recently, 3D Gaussian splatting (3DGS) \cite{kerbl3dgs} has emerged as an \changed{explicit} alternative for representing radiance fields, achieving state-of-the-art (SoTA) rendering quality in novel view synthesis (NVS) while enabling real-time rasterization. Subsequent works in the domain gracefully inherit its computational advantage and adapt the method to various vision-based tasks, including surface reconstruction and scene decomposition from multi-view images. Despite producing high-quality meshes, SoTA \changed{explicit} surface reconstruction methods often struggle to recover objects exhibiting reflection. In particular, they either rely solely on view-dependent radiance functions, i.e, spherical harmonics \cite{Dai2024GaussianSurfels, Huang2DGS2024, Yu2024GOF, guedon2025milo}, incorporate a small multi-layer perceptron (MLP) model for exposure compensation \cite{Chen2025PGSR}, or normal priors from pretrained models \cite{wang2024GausSurf}. These limited appearance modelings may perform well for scenes with diffuse surfaces, yet suffer from those exhibiting highly varying photometric details, as shown in Figure \ref{fig-sota-reflective}. \changed{Conversely, 3DGS-based} methods with sophisticated appearance modeling excel in decomposing material properties, yet little work has been demonstrated for their capability of \changed{complementing mesh reconstruction tasks for reflective surfaces}. Recent SoTA \changed{explicit} methods in material decomposition \changed{realized the significance of the underlying surface geometry and and brought these tasks together} by borrowing components from neural-based approaches such as SDF backbones \cite{zhu2025gsror2}, geometry segmentation priors \cite{glossygs}, or tensor factorization \cite{Zhang2025refgs} \changed{to maintain high-fidelity surface geometry}. \changed{Despite delivering faithful decomposition results with geometrically feasible meshes}, they inherit the performance penalty carried over by said neural components, hindering their \changed{runtime} at scale. Moreover, these methods still demonstrate limited capability of reconstructing surfaces with sharp edges and geometric details, making them unfavorable for recovering diffuse but highly intricate objects.

To this end, we propose a \changed{material-aware optimization framework for high-fidelity mesh reconstruction from multi-view images based on
3D Gaussian Splatting}, referred to as GS-2M. Our goal is to derive a solution that maintains the reconstruction quality to be on par with current SoTA surface reconstruction methods while simultaneously handling reflective objects, as demonstrated in Figure \ref{fig-teaser}. By incorporating material parameters into the training pipeline, we identify the appearance properties of the underlying object and eliminate geometric artifacts caused by view-dependent effects, producing watertight meshes and smoother surfaces. To further eliminate neural components in this joint optimization framework, we propose a novel roughness supervision strategy based on multi-view photometric variation, completely independent of priors from pre-trained models. This allows us to scale the runtime performance of our framework without being constrained by external factors. Finally, we validate our method with the DTU \cite{jensen2014dtu} and TanksAndTemples (TnT) \cite{Knapitsch2017tnt} benchmarks to demonstrate its ability to maintain reconstruction performance, and the Shiny Blender Synthetic dataset \cite{Verbin2022refnerf} for qualitative comparisons with SoTA surface reconstruction methods. To summarize, our contributions are:

\begin{itemize}
    \item A \changed{novel} framework jointly optimizing 3D Gaussians for both mesh reconstruction and material decomposition from multi-view images, delivering comparable mesh reconstruction quality to SoTA, while being resilient to reflective surfaces.
    \item A roughness supervision strategy based on multi-view photometric variation, eliminating the dependence on neural components for appearance modeling.
    \item An integration of occlusion-aware check and multi-view normal consistency for SoTA mesh reconstruction methods, improving NVS performance while maintaining the reconstruction quality.
\end{itemize}

%------------------------------------------------------------------------
\section{Related Work}
\changed{In this section}, we review recent methods based on 3DGS for mesh reconstruction and material decomposition from multi-view \changed{posed} images. For completeness, classical and neural implicit reconstruction approaches are also discussed in Section \ref{sec-review-traditional} and \ref{sec-review-neural}, respectively. Section \ref{sec-review-3dgs} then extensively reviews the recent SoTA neural explicit (3DGS-based) works that inspired GS-2M’s design.

\subsection{Traditional methods}\label{sec-review-traditional}
Reconstruction of 3D geometry from photographs is an ill-posed problem that relies on certain assumptions about the underlying scene \cite{Furukawa2015}. Among these cues, stereo correspondence has exhibited reasonable robustness and propelled a class of reconstruction algorithms known as multi-view stereo (MVS) \cite{Seitz2006MVS}. MVS shares largely the same principle as GS-2M and other 3DGS-based methods: reconstruct the 3D geometry of the scene from multi-view images. Structure-from-Motion (SfM) \cite{Snavely2006sfm} is widely adopted to recover \changed{camera intrinsics and poses} for each image, with an initial sparse pointcloud as a byproduct. From there, \changed{3D meshes} can be reconstructed via voxel-based optimization \cite{Seitz1990voxel, Sinha2007graphcut}, feature point growing \cite{Furukawa2010, Wu2010quasi}, or depth fusion \cite{schonberger2016, Xu2018depth, Huang2024visbility}. Like other traditional methods, their performance is affected by inconsistent appearance across input images, making it challenging to accurately capture complete geometric representations due to the ambiguities in the correspondence. By contrast, we model the appearance properties of the target object with a handful of material parameters as part of our pipeline, making it resilient to abrupt photometric changes across views.

\subsection{Neural implicit methods}\label{sec-review-neural}
As NeRF is not mainly designed for surface reconstruction, radiance fields are
replaced with implicit surfaces \cite{Niemeyer2020, Oechsle2021unisurf} or signed distance functions (SDFs) \cite{yariv2020} to better define isosurfaces from a volume density. These representations are later reparameterized \cite{wang2021neus, yariv2021volume} to be compatible with neural volume rendering as employed by NeRF. To further increase the reconstruction quality, auxiliary information is baked into the training pipeline as priors, such as patch warping with co-visibility masks \cite{darmon2022neuralwarp}, sparse SfM points \cite{Fu2022, Zhang2022regsdf}, semantic segmentation \cite{Guo2022}, monocular depth \cite{Sun2022}, or monocular geometric features \cite{Yu2022monosdf}. On the other hand, Neuralangelo \cite{li2023neuralangelo} leverages hash encodings as introduced by InstantNGP \cite{mueller2022instant} to eliminate the need for auxiliary guidance, while HF-NeuS \cite{Wang2022hfneus} adopts coarse-to-fine optimization for improved surface details. As previously stated, these implicit methods are SoTA in mesh reconstruction but demand expensive training resources, and attempts have been made to speed up their training time \cite{wang2023neus2, yariv2023bakedsdf, Li2024voxsurf}. However, these enhancements often come at the expense of reconstruction quality, while \changed{neural explicit} methods retain sufficient performance without requiring high computing resources.

Neural implicit approaches for material decomposition often stem from inverse rendering frameworks. Early methods only consider direct lighting \cite{Boss2021nerd, Boss2021neuralpil, Zhang2021physg} to trade quality for computation speed. Subsequent works \cite{Srinivasan2021nerv, Yao2022neilf, Jin2023tensoir} introduce indirect lighting but produce inferior results on highly reflective objects. To combat this, Ref-NeRF \cite{Verbin2022refnerf}  decomposes colors into diffuse and specular terms, while NeRO \cite{liu2023nero} designs a novel light representation based on the split-sum approximation \cite{karis2013realshading} of the rendering equation. Most recently, TensoSDF \cite{li2024tensosdf} achieves SoTA decomposition performance by expressing geometry as an implicit SDF and incorporating roughness-aware radiance fields.

% and employs normal priors from StableNormal \cite{Ye2024stablenormal} to further supervise depth-normal consistency
% This helps us maintain the reconstruction performance with SoTA methods while simultaneously handling reflective surfaces.

\subsection{Neural explicit methods}\label{sec-review-3dgs}
SuGaR \cite{guedon2023sugar} was among the first to adapt 3DGS for mesh reconstruction. By minimizing the SDF derived from the Gaussians and the scene density function, they encourage points to better align with the underlying surface. However, SuGaR only approximates the planarity of Gaussians with their minimum scaling axes, resulting in insufficient constraints on their overall shape during training. To address this loose approximation, 2D Gaussian Splatting (2DGS) \cite{Huang2DGS2024} and GaussianSurfels \cite{Dai2024GaussianSurfels} replace 3D Gaussian points with planar primitives. In particular, they collapse 3D Gaussians to planar ellipses to maximize the alignment between points and the object surface. Despite achieving view-consistent geometry, 2DGS and GaussianSurfels rely on mean/median or blended z-depth in camera space, which may cause ambiguity or biased depth rendering. At the other extreme, Gaussian Opacity Fields (GOF) \cite{Yu2024GOF} constructs an opacity field from the trained Gaussians to extract the underlying surface by identifying its level set, drawing inspiration from ray-traced volume rendering of 3D Gaussians. MILo \cite{guedon2025milo} later incorporates the mesh extraction step into the optimization process of GOF, simultaneously minimizing the image-based and surface consistency losses. However, GOF and MILo suffer from the expensive computational resources required for their sophisticated mesh extraction and training algorithms. PGSR \cite{Chen2025PGSR} and GausSurf \cite{wang2024GausSurf} are \changed{recent SoTA explicit methods} in the field, with the former introducing unbiased depth rendering and multi-view constraints imposed for geometric and photometric consistency, while the latter borrowing the patch-matching technique from MVS to refine the rendered depth across views. As stated before, they both struggle to reconstruct reflective objects due to the lack of appearance modeling, producing distorted meshes even for simple scenes.

% Consequently, we build our method on top of PGSR to inherit its multi-view construct and introduce shading functions to capture material properties and scene lighting.

To address view-dependent issues stemming from highly specular surfaces, \cite{Jiang2024GaussianShader} introduces simplified shading functions to capture light-surface interactions, while \cite{Ye2024Deferred3DGS} employs deferred rendering with a trainable reflection strength parameter. These ideas are quickly adopted in the literature and further enhanced for scene relighting \cite{Bi2024gs3, Wu2025DeferredGS, Shi2025GIR} and inverse rendering frameworks \cite{ye2024geosplatting, Liang2024gsir, zhou2025rtrgs}. To further factorize lighting, most works employ differential environment cubemaps \cite{Laine2020modular} and sample them during training at various mip levels. This helps separate lighting from the object’s intrinsic appearance, allowing for more accurate material decomposition. GS-ROR$^2$ \cite{zhu2025gsror2}, GlossyGS \cite{glossygs}, and Ref-GS \cite{Zhang2025refgs} are SoTA works in this context by also realizing the significance of the underlying geometry to the decomposition performance. In particular, GS-ROR$^2$ jointly optimizes an SDF backbone for robust geometry, GlossyGS incorporates micro-facet features and priors to supervise neural materials, and Ref-GS builds upon 2DGS with tensorial factorization for material decomposition.

\section{Method}
To set up the necessary terminology and notations, we provide a prelude to 3DGS in Section \ref{sec-med-pre}. Section \ref{sec-med-gs2m} discusses in detail the construct of GS-2M, including the choice of point primitive, definitions of depth and normals, and augmented material parameters.

\subsection{Preliminary}\label{sec-med-pre}
To represent a scene, \cite{kerbl3dgs} initializes $n$ anisotropic 3D Gaussians, $\{\mathcal{G}_0,\ \mathcal{G}_1,\ \mathcal{G}_2,\ \cdots,\ \mathcal{G}_{n-1}\}$, where each Gaussian $\mathcal{G}_i$ is parameterized by a center $\mu_i \in \mathbb{R}^3$, a scaling vector $\mathbf{s}_i \in \mathbb{R}^3$, and a quaternion $\mathbf{q}_i \in \mathbb{R}^4$. These are learnable parameters, with $\mathbf{s}_i$ and $\mathbf{q}_i$ further defining a scaling matrix $S_i \in \mathbb{R}^{3 \times 3}$ and a rotation matrix $R_i \in \mathrm{SO}(3)$. Evaluating $\mathcal{G}_i$ at a position $\mathbf{x} \in \mathbb{R}^3$ thus follows $\mathcal{G}_i(\mathbf{x}) = e^{-\frac{1}{2}(\mathbf{x} - \mu_i)^\top \Sigma_i^{-1}(\mathbf{x} - \mu_i)}$, where $\Sigma_i = R_i S_i S^\top_i R^\top_i$ is the 3D covariance matrix of $\mathcal{G}_i$. Suppose there's a camera $\mathcal{C}$ in the scene whose viewing transform is $V \in \mathbb{R}^{4 \times 4}$, we can apply EWA volume splatting \cite{Zwicker2001} to find the 2D projection $\mathcal{G}^{2\mathrm{D}}_i$ of $\mathcal{G}_i$ as seen by $\mathcal{C}$. Given the projected 2D Gaussians in image space, the alpha value $\alpha_i$ of any $\mathcal{G}^{2\mathrm{D}}_i$ can be evaluated for any pixel $\mathbf{p} \in \mathbb{R}^2$ by evaluating $\mathcal{G}^{2\mathrm{D}}_i(\mathbf{p})$, multiplied with a learnable opacity value $\phi_i$ corresponding to $\mathcal{G}_i$. From there, Equation \ref{eq-alpha-blending} formulates the $\alpha$-blending process to compute the rendered image $\hat{\mathcal{I}}$, where per pixel linear color $\hat{\mathcal{I}}(\mathbf{p}) \in \mathbb{R}^3$ is the alpha-composite value of the view-dependent color $\mathbf{c}_i \in \mathbb{R}^3$ evaluated from $\mathcal{G}_i$'s learnable spherical harmonics (SH) coefficients. Note that $i$ in the equation indexes over \changed{the set $G_\mathbf{p}$ of} 2D splats that contribute meaningful values to pixel $\mathbf{p}$ in front-to-back order.

\begin{equation}\label{eq-alpha-blending}
    \hat{\mathcal{I}}(\mathbf{p}) = \sum_{i\ 
\in\ G_\mathbf{p}} T_i \alpha_i \mathbf{c}_i,\ \ T_i = \prod^{i - 1}_{j=0}(1 - \alpha_j)
\end{equation}

Since most learnable parameters are randomly initialized, $\hat{\mathcal{I}}$ is not yet similar to the ground-truth image $\mathcal{I}$ captured under the same viewpoint. A loss function $\mathcal{L}$ can therefore be defined to measure how far $\hat{\mathcal{I}}$ is to $\mathcal{I}$, and a training process is then established to minimize $\mathcal{L}$ by optimizing all learnable parameters contributing to the rasterization of $\hat{\mathcal{I}}$. 3DGS uses a simple RGB photometric loss $\mathcal{L}_\mathrm{rgb}$ composed of pixel-wise mean absolute error between $\hat{\mathcal{I}}$ and $\mathcal{I}$, weighted sum with an SSIM term, detailed in the supplementary.

% \cite{kerbl3dgs} uses a simple RGB photometric loss for $\mathcal{L}$, as described in Equation \ref{eq-Lrgb-loss}. Here, $L_1$ is the pixel-wise mean absolute error between $\hat{\mathcal{I}}$ and $\mathcal{I}$, while $L_\mathrm{D-SSIM}$ measures their structural similarity index following \cite{Baker2024DSSIM}. $\lambda_\mathrm{SSIM}$ in the equation is set to $0.2$, identical to 3DGS and previous works.

% \begin{equation}\label{eq-Lrgb-loss}
%     \mathcal{L}_\mathrm{rgb} = (1 - \lambda_\mathrm{SSIM})L_1 + \lambda_\mathrm{SSIM}L_\mathrm{D-SSIM}
% \end{equation}

\subsection{GS-2M}\label{sec-med-gs2m}
As an overview, we construct GS-2M from PGSR \cite{Chen2025PGSR} to maintain SoTA reconstruction performance. In particular, we employ unbiased depth rendering \cite{Chen2025PGSR, Huang2024visbility} and multi-view constraints \cite{Chen2025PGSR, Huang2024visbility, wang2024GausSurf} to encourage high-fidelity mesh reconstruction. For material decomposition, we draw inspiration from \cite{Jiang2024GaussianShader, Liang2024gsir, zhu2025gsror2} and introduce two additional per-Gaussian learnable parameters with lighting captured via differential environment cubemaps. As stated before, we propose a novel roughness supervision strategy to help supervise material parameters without relying on neural components. In short, we leverage the existing multi-view construct of PGSR and measure photometric variation across views using the normalized cross correlation (NCC) loss applied to warped patches.

\subsubsection*{Unbiased depth rendering and normal as shortest axis}

\begin{figure}
    \centering
    \includegraphics[width=1\linewidth]{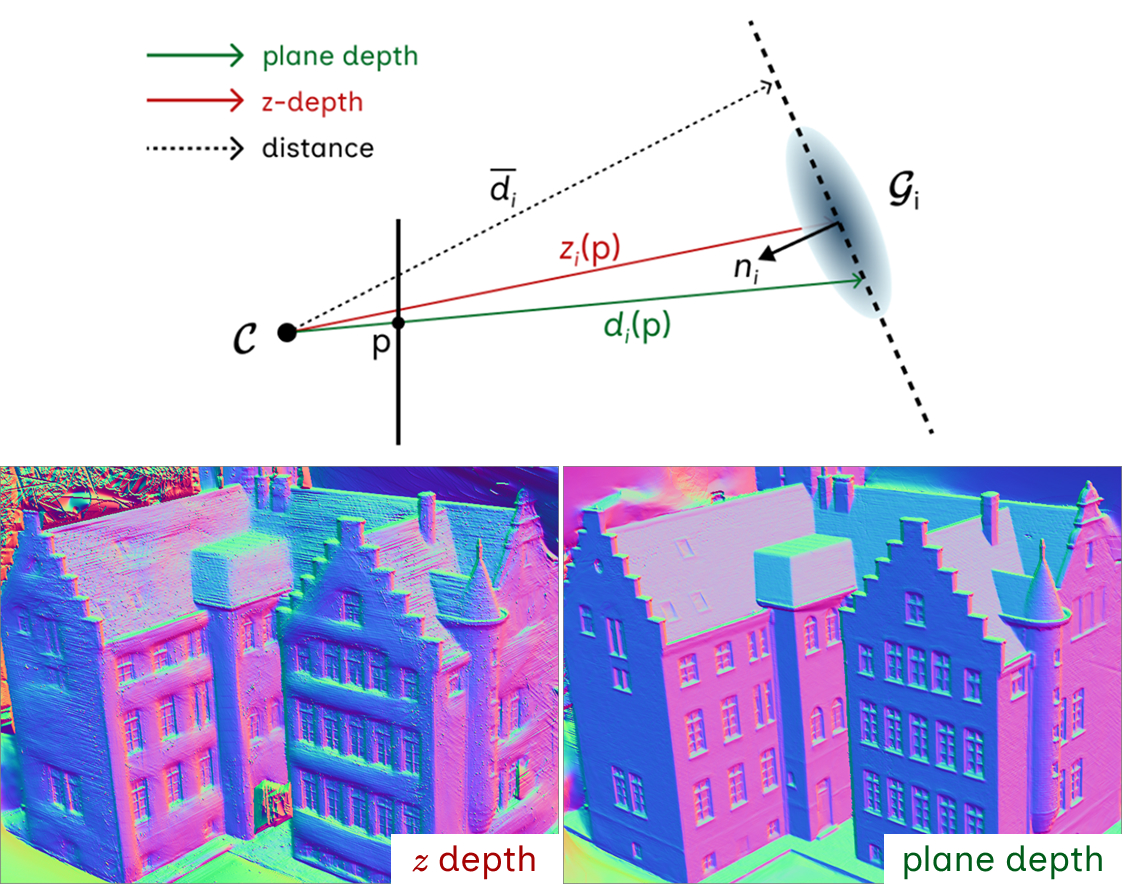}
    \caption{Comparing depth quality when training with $z$-depth (left) and plane depth (right). We compare normal maps derived from rendered depth maps via a Sobel-like operator for better visualization. Training with $z$-depth results in biased and noisy depth values, while plane depth allows for more accurate and consistent distributions of Gaussians to capture geometric details.}
    \label{fig-unbiased-depth}
\end{figure}

As highlighted in recent works \cite{Huang2024visbility, Chen2025PGSR}, blending $z$-depth in camera space results in biased depth. Concretely, nearby pixels observing the same Gaussian $\mathcal{G}_i$ have similar $z$ values even though their true depth values are different, as illustrated in Figure \ref{fig-unbiased-depth}. The unbiased depth is instead identified based on the hypothetical plane perpendicular to the normal $\mathbf{n}_i \in \mathbb{R}^3$ of $\mathcal{G}_i$. Following previous works \cite{Ye2024Deferred3DGS, Jiang2024GaussianShader, Liang2024gsir, Chen2025PGSR, zhu2025gsror2}, we assign the orientation axis (a column vector of $R_i$) corresponding to the shortest scaling direction (a scalar in $\mathbf{s}_i$) as $\mathbf{n}_i$, normalized and flipped to face $\mathcal{C}$. From there, we can compute the distance value $\bar{d}_i = \mu'_i \cdot (R_i\mathbf{n}_i)$ from $C$ to $\mathcal{G}_i$, where $\mu'_i$ is the mean $\mu_i$ of $\mathcal{G}_i$ transformed to camera space. By replacing $\mathbf{c}_i$ in Equation \ref{eq-alpha-blending} with $\bar{d}_i$ and $\mathbf{n}_i$, respectively, we obtain the distance map $\bar{\mathcal{D}}$ and normal map $\mathcal{N}$ defined for all $\mathbf{p}$. The unbiased depth map $\mathcal{D}$ is finally computed by rescaling the distance map with the inverted cosine of the angle between the normal and ray directions. Equation \ref{eq-plane-depth} details this operation, where $\tilde{\mathbf{p}}$ is the homogeneous version of $\mathbf{p}$, i.e $\tilde{\mathbf{p}} = [\mathbf{p}, 1]^\top$, and $K \in \mathbb{R}^{3\times 3}$ is the intrinsic of $\mathcal{C}$.

\begin{equation}\label{eq-plane-depth}
    \mathcal{D}(\mathbf{p}) = \frac{\bar{\mathcal{D}}(\mathbf{p})}{\mathcal{N}(\mathbf{p}) \cdot (K^{-1}\tilde{\mathbf{p}})}
\end{equation}

The use of plane depth requires planar or near-planar Gaussians, for which we employ the plane loss term $\mathcal{L}_\mathrm{plane}$ to penalize one of the scaling scalars of all $\mathbf{s}_i$. We further impose the depth-normal constraint $\mathcal{L}_\mathrm{dn}$ as used in previous works \cite{Huang2DGS2024, Huang2024visbility, Dai2024GaussianSurfels, Yu2024GOF, Chen2025PGSR, wang2024GausSurf, Jiang2024GaussianShader, Liang2024gsir, zhu2025gsror2} to encourage their consistency. \changed{Both are defined in Equation \ref{eq-Lplane-n-Ldn} and further} detailed in the supplementary.

\begin{equation}\label{eq-Lplane-n-Ldn}
    \mathcal{L}_\mathrm{plane} = \frac{\lambda_\mathrm{plane}}{|\mathcal{V}|}\sum_\mathbf{s}||\min(\mathbf{s}_i)||_1,\ \ \mathcal{L}_\mathrm{dn} = \frac{\lambda_\mathrm{dn}}{|\mathcal{N}|}\sum_\mathbf{p} |\nabla\mathcal{I}(\mathbf{p})|^2 ||\mathbf{n}_\mathbf{p} - \hat{\mathbf{n}}_\mathbf{p}||_1
\end{equation}

\subsection*{Multi-view normal consistency and occlusion-aware filtering}\label{sec-mv-new}
\begin{figure}
    \centering
    \includegraphics[width=1\linewidth]{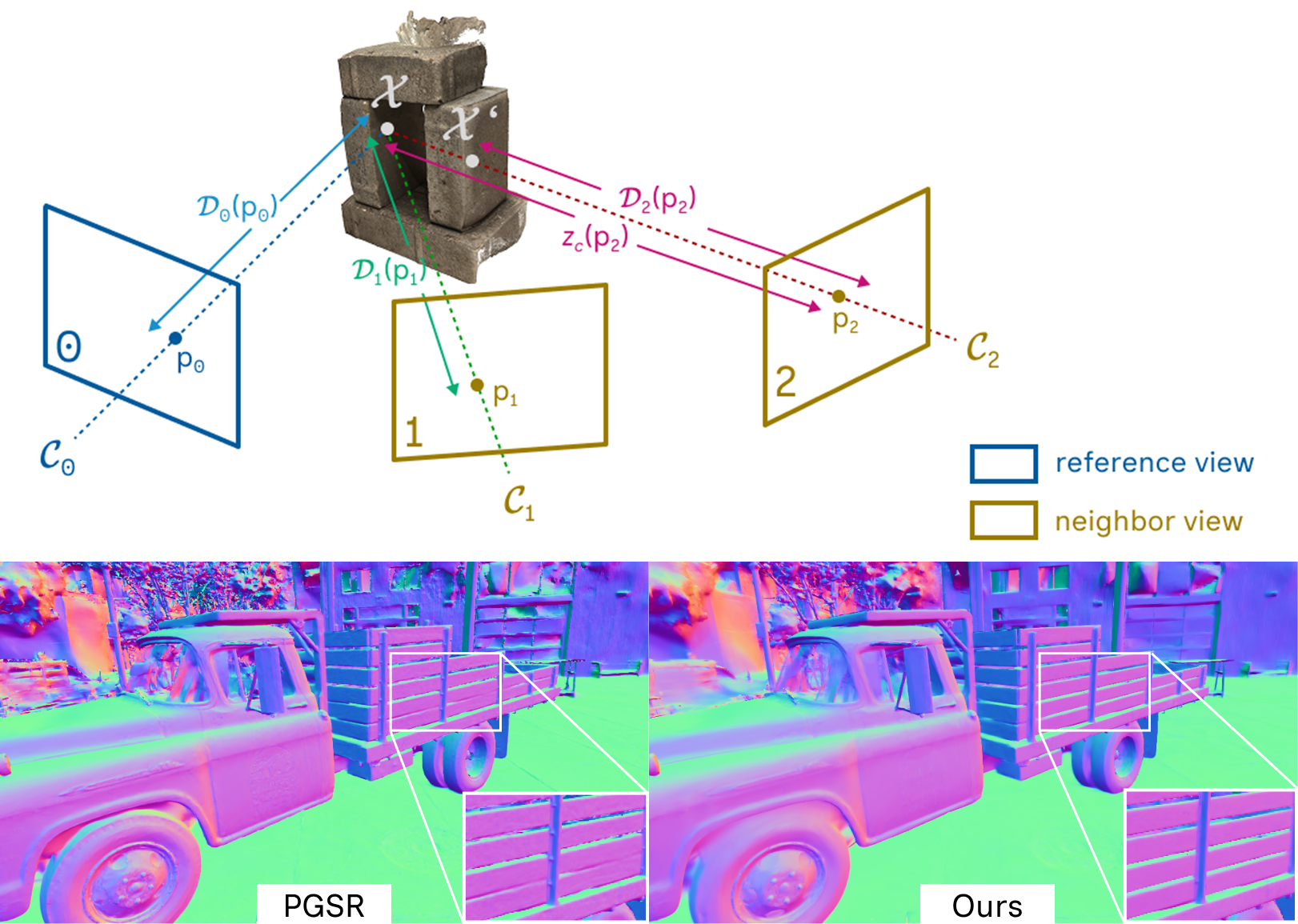}
    \caption{Filtering invalid correspondences in neighboring views (top) and enhanced multi-view constraints with normal consistency (bottom). Correspondence pixel $\mathbf{p}_2$ of neighbor view $\mathcal{C}_2$ is excluded from multi-view loss calculations because its depth value is less than the $z$-coordinate of the camera-space point $\mathcal{X}$ back-projected from $\mathbf{p}_0$ in the reference view. We also sample values from normal maps rendered at reference and neighboring views to provoke multi-view normal consistency in high-frequency regions.}
    \label{fig-mv-pgsr}
\end{figure}

As stated previously, we adopt PGSR \cite{Chen2025PGSR}'s multi-view loss $\mathcal{L}_\mathrm{mv}$---a weighted sum of multi-view geometric $L_\mathrm{g}$ and multi-view photometric $L_\mathrm{p}$ terms. However, PGSR only minimizes the reprojection error between sampled pixels in the reference view and those found via forward--backward projection of a neighboring view, resulting in underconstrained supervision for $L_\mathrm{g}$. Inspired by \cite{Huang2024visbility}, we further encourage multi-view normal consistency by minimizing the difference in normal directions between the reference and neighboring views at sampled points in world space. This helps us achieve more consistent geometry in regions with high-frequency textures, as demonstrated in Figure \ref{fig-mv-pgsr}. Please refer to the supplementary for the modified $L_\mathrm{g}$ term and details of $\mathcal{L}_\mathrm{mv}$.

% Instead of empirically rejecting correspondence pixels with large reprojection errors, we filter invalid correspondences based on rendered depth maps of neighboring views and $z$-depth of back-projected points in camera space.

% In other words, $\mathcal{X}$ is occluded by $\mathcal{X}'$ when observed from $\mathcal{C}_2$, making geometric constraints applied on these correspondences invalid.

% As a result, we achieve more consistent geometric details in regions with high-frequency textures.

Another improvement we apply on top of PGSR is robust occlusion-aware filtering, as introduced in \cite{Huang2024visbility}. Specifically, PGSR only approximates invalid correspondences by thresholding large reprojection noises, which is unreliable and ambiguous. We instead explicitly detect and reject these invalid pixels by comparing their depth values in the neighbor view’s rendered depth map with $z$-coordinates of back-projected points in the same camera space, as shown in Figure \ref{fig-mv-pgsr}.

% Figure \ref{fig-mv-pgsr} depicts this robust occlusion check, where pixels in neighboring views with depth values less than their corresponding camera $z$-depth are rejected from multi-view consideration.

\subsubsection*{Material modeling and differential environment lighting}
We introduce two additional per-Gaussian learnable parameters---albedo $\mathbf{a}_i \in \mathbb{R}^3$ and roughness $\rho_i \in [0,1]$---to decompose intrinsic properties of surface appearance, allowing the optimization process to treat smooth (reflective) and rough (diffuse) regions differently. To learn these parameters, we employ a physically based rendering (PBR) pipeline, where the BRDF is formulated with the Cook-Torrance microfacet model \cite{Cook1982, Torrance2007}. In essence, the shading model combines diffuse and specular reflection, with the specular component governed by microfacet theory. We provide the mathematical construct of the BRDF in the supplementary.

Similar to depth and normal maps, we also render the albedo map $\mathcal{A}$ and roughness map $\mathcal{R}$ by $\alpha$-blending per-Gaussian albedo and roughness parameters. Additionally, the Cook-Torrance shading model depends on a metallic fraction value between $0$ and $1$, for which we approximate from the roughness map as $\mathcal{M} = 1 - \mathcal{R}$. Finally, we composite all G-buffer renders into a PBR image $\bar{\mathcal{I}}$ (not to be confused with the rasterized image $\hat{\mathcal{I}}$) via a deferred rendering step \cite{Ye2024Deferred3DGS, Wu2025DeferredGS}. \changed{Concretely, we perform deferred physically based shading in image space by feeding the rendered G-buffer channels (albedo, roughness, depth, and normals) into a Cook--Torrance BRDF, from which we separately compute the diffuse and specular lighting components before compositing them into the final PBR image.} The RGB photometric loss can therefore be replaced with the PBR photometric loss $\mathcal{L}_\mathrm{pbr}$ to supervise all components contributing to the final PBR composite. Similar to $\mathcal{L}_\mathrm{rgb}$, $\mathcal{L}_\mathrm{pbr}$ consists of the pixel-wise $L_1$ term and the SSIM term, applied on $\bar{\mathcal{I}}$ and $\mathcal{I}$, \changed{as defined in Equation \ref{eq-Lpbr}}.

\begin{equation}\label{eq-Lpbr}
    \mathcal{L}_\mathrm{pbr} = (1 - \lambda_\mathrm{SSIM})L_1 + \lambda_\mathrm{SSIM}L_\mathrm{D-SSIM}
\end{equation}

% In essence, the BRDF can be split into a diffuse and specular component for the computation of the rendering equation

% In particular, it approximates specular shading as the combination of a constant roughness-aware lighting term and a BRDF's response term.

Per \cite{Kajiya1986}'s rendering equation, the PBR compositing of the G-buffer requires incoming irradiance of the environment lighting. As with previous works \cite{Liang2024gsir, zhu2025gsror2}, we employ a differential environment cubemap \cite{Laine2020diffrast} to learn incoming radiance, and prefilter it at various mip levels to sample irradiance values at runtime. As diffuse shading is independent of the viewing direction, we sample prefiltered lighting values from the base level of the cubemap. For the specular component, however, we adopt \cite{karis2013realshading}'s split-sum approximation to help with the computation of the view-dependent integral, combining a constant roughness-aware lighting term and a BRDF's response term. The former is similar to irradiance sampled from a prefiltered cubemap, but at the mip level corresponding to the supplied roughness value. The latter, on the other hand, is pre-computed and stored in a 2D lookup texture (LUT) for fast retrieval of BRDF's responses at runtime. Please refer to the supplementary for the mathematical constructs of the rendering equation and lighting computation in our PBR pipeline.

% Despite faithfully addressing the problem, they inherit the performance penalty carried over by external factors.

\subsubsection*{Multi-view roughness supervision}
Training with $\mathcal{L}_\mathrm{pbr}$ alone to supervise appearance parameters is extremely under-constrained, resulting in noisy lighting and incoherent decomposition results. Previous works \cite{Zhang2025refgs, glossygs, zhu2025gsror2} thus exploit neural components in terms of encoders--decoders or pretrained priors to learn scene parameters. We instead propose a lightweight roughness supervision strategy based solely on multi-view photometric variation, eliminating the need for neural modeling of scene components.

%  We detect rough and smooth regions by comparing grayscale patches of the reference view with their warped counterparts in a nearby view.
% By hoisting $\lambda_\mathrm{ref}$ as a hyperparameter, we allow the optimization process to opt for a suitable threshold based on the nature of the reflection.

\begin{figure*}
    \centering
    \includegraphics[width=1\textwidth]{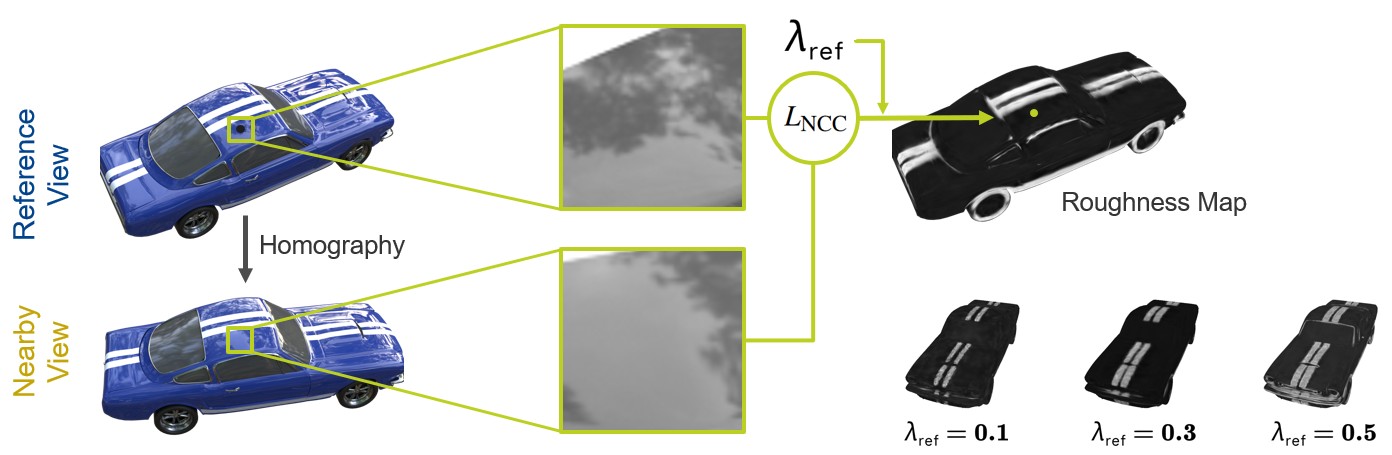}
    \caption{Roughness supervision based on multi-view photometric variation. The photometric variation is quantified with the NCC error $L_\mathrm{NCC}$, and a thresholding value $\lambda_\mathrm{ref}$ is used to penalize or reward the corresponding roughness values sampled from the rendered roughness map of the reference view. As $\lambda_\mathrm{ref}$ increases (bottom right), more and more regions become diffuse, i.e., their multi-view photometric variation is not regarded as inconsistency caused by reflective surfaces.}
    \label{fig-mv-rough}
\end{figure*}

As stated earlier, we recognize the multi-view photometric NCC error between image patches as an indicator for highly reflective regions. If the surface nature is strongly view-dependent, switching to nearby viewpoints will likely cause the corresponding warped patch's texture to change significantly. We measure this variation for $3 \times 3$ grayscale patches, constructed from sampled pixels of ground-truth images, illustrated in Figure \ref{fig-mv-rough}. The use of NCC helps emphasize texture-based similarity rather than absolute pixel intensity, thereby making the error resilient to brightness changes or geometric misalignment. Equation \ref{eq-ncc} describes how $L_\mathrm{NCC}$ is computed given a sampled image pixel $\mathbf{p}$, where $\mathcal{P}_\mathrm{r}$ is the $3 \times 3$ image patch enclosing $\mathbf{p}$ in the reference view, and $\hat{\mathcal{P}}_\mathrm{n}$ is the warped version of $\mathcal{P}_\mathrm{r}$ in a nearby view via homography. \changed{We maintain, for each ground-truth image, a heuristic list of nearby viewpoints that are close to the reference viewpoint by jointly thresholding the Euclidean distance between camera centers and the angular difference between viewing rays, and then selecting a fixed number of candidates via stratified sampling over the sorted list.}

\begin{equation}\label{eq-ncc}
    L_{\mathrm{NCC}}(\mathbf{p}) = 1 - \frac{\sum_{p} ( \mathcal{P}_\mathrm{r}(p) - \mu_{\mathcal{P}_{\mathrm{r}}} )  (\hat{\mathcal{P}}_\mathrm{n}(p) - \mu_{\hat{\mathcal{P}}_{\mathrm{n}}})}
{\sqrt{\sum_{p} \left( \mathcal{P}_\mathrm{r}(p) - \mu_{\mathcal{P}_{\mathrm{r}}} \right)^2} \sqrt{\sum_{p} ( \hat{\mathcal{P}}_\mathrm{n}(p) - \mu_{\hat{\mathcal{P}}_{\mathrm{n}}} )^2}}
\end{equation}

% Once $L_\mathrm{NCC}$ is calculated for every sampled pixel $\mathbf{p}$, we penalize or reward the roughness values at the corresponding pixels in the rendered roughness map, using a threshold $\lambda_\mathrm{ref}$. Specifically, roughness values at pixels whose $L_\mathrm{NCC}$ exceeds $\lambda_\mathrm{ref}$ are added to the roughness loss $\mathcal{L}_\mathrm{ro}$, while those less than the same threshold are subtracted from $\mathcal{L}_\mathrm{ro}$. Finding a fixed $\lambda_\mathrm{ref}$ value to work for all scenes is not plausible since reflective clues manifest in various ways. For example, some objects may exhibit very high-frequency reflection images and thus need a higher $\lambda_\mathrm{ref}$ to maintain non-diffuse regions. By contrast, objects with blurred and fuzzy reflection clues require a lower $\lambda_\mathrm{ref}$ to make slight changes in multi-view photometric more sensitive for $\mathcal{L}_\mathrm{ro}$. As a result, we leave $\lambda_\mathrm{ref}$ as one of the optimization parameters that can be tuned depending on the nature of the scene, as shown in Figure \ref{fig-mv-rough}.

Once $L_\mathrm{NCC}$ is computed for each sampled pixel $\mathbf{p}$, we modulate the corresponding roughness values using a threshold $\lambda_\mathrm{ref}$: roughness values at pixels with $L_\mathrm{NCC} > \lambda_\mathrm{ref}$ are added to the roughness loss $\mathcal{L}_\mathrm{ro}$, and those below the threshold are subtracted. A fixed $\lambda\mathrm{ref}$ is not suitable for all scenes, as sharp reflections require higher thresholds to preserve non-diffuse regions, whereas blurred reflections benefit from lower thresholds to increase the sensitivity of $\mathcal{L}_\mathrm{ro}$ to weak multi-view cues. \changed{In practice, we recommend choosing $\lambda_\mathrm{ref}$ based on the dominant material characteristics of the scene: for scenes containing primarily glossy or mirror-like objects with sharp view-dependent highlights, a higher threshold (e.g., $\lambda_\mathrm{ref} \in [0.9, 1.1]$) helps avoid over-smoothing specular regions, whereas for scenes dominated by semi-glossy or rough materials with weak or blurred reflections, a lower threshold (e.g., $\lambda_\mathrm{ref} \in [0.6, 0.9]$) makes the loss more responsive to subtle multi-view photometric inconsistencies. For mixed-material scenes, we found that intermediate values around $\lambda_\mathrm{ref} \approx 0.9$ provide a robust trade-off without requiring per-object tuning.}

% In practice, we set $\lambda_\mathrm{ref} < 1.0$ when recovering reflective objects, and to $1.0$ otherwise. 

%  Notice the white stripes on the car's surface are not reflective.

\begin{figure}
    \centering
    \includegraphics[width=1\linewidth]{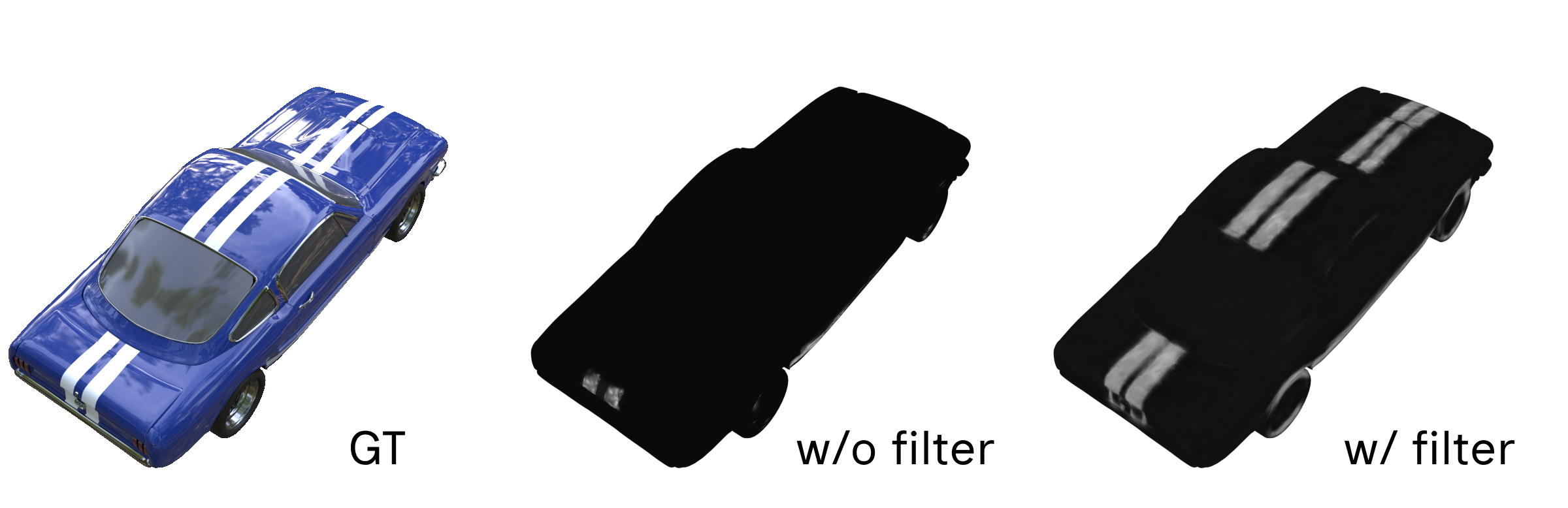}
    \caption{Simply relying on $L_\mathrm{NCC}$ to identify multi-view photometric variation results in incorrect roughness supervision (middle) at textureless regions. Replacing these regions with gradient-based patches helps $L_\mathrm{NCC}$ produce more faithful results (right).}
    \label{fig-ncc-filter}
\end{figure}

There's a limitation to the described supervision strategy so far, in which textureless regions yield high $L_\mathrm{NCC}$ despite being non-reflective. When there is no variation, the denominator of Equation \ref{eq-ncc} approaches zero, making $L_\mathrm{NCC}$ explode and unstable. To combat this, we filter out pixels in the reference views where the standard deviation $\sigma_r(\mathbf{p}) = \sqrt{\sum_{p} (\mathcal{P}_{\mathrm{r}}(p) - \mu_{\mathcal{P}_{\mathrm{r}}})^2}$ less than an empirically derived threshold of $0.01$, and replace them with $L_{\mathrm{NCC}}$ but applied on the gradient versions of $\mathcal{P}_\mathrm{r}$ and $\hat{\mathcal{P}}_\mathrm{n}$, as illustrated in Figure \ref{fig-ncc-filter}. Once $L_\mathrm{NCC}$ values are filtered, we implement the roughness loss $\mathcal{L}_\mathrm{ro}$ using Equation \ref{eq-loss-ro}. Here, to avoid abrupt constraining of roughness values $\mathcal{R}(\mathbf{p})$, we remap $L_\mathrm{NCC}$ via the $\tanh$ function, and apply a scale factor $k_\mathrm{ro}$ that controls how rapidly the weights change from promoting to penalizing $\mathcal{R}(\mathbf{p})$. We set $k_\mathrm{ro} = 8.0$, determined via empirical experiments.

\begin{table*}[t]
\centering
\resizebox{\textwidth}{!}{
\begin{tabular}{lcccccccccccccccc|c}
\toprule
\textbf{CD} $\downarrow$ & 24 & 37 & 40 & 55 & 63 & 65 & 69 & 83 & 97 & 105 & 106 & 110 & 114 & 118 & 122 & Mean & Time \\
\midrule
VolSDF & 1.14 & 1.26 & 0.81 & 0.49 & 1.25 & 0.70 & 0.72 & 1.29 & 1.18 & 0.70 & 0.66 & 1.08 & 0.42 & 0.61 & 0.55 & \textbf{0.86} & $\sim$ 12h \\
NeuS & 0.83 & 0.98 & 0.56 & 0.37 & 1.13 & 0.59 & 0.60 & 1.45 & 0.95 & 0.78 & 0.52 & 1.43 & 0.36 & 0.45 & 0.45 & \textbf{0.77} & $\sim$ 8h \\
RegSDF & 0.60 & 1.41 & 0.64 & 0.43 & 1.34 & 0.62 & 0.60 & 0.90 & 0.92 & 1.02 & 0.60 & 0.59 & \cellcolor{best}0.30 & 0.41 & 0.39 & \textbf{0.72} & $\sim$ 3.5h \\
NeuS2 & 0.56 & 0.76 & 0.49 & 0.37 & 0.92 & 0.71 & 0.76 & 1.22 & 1.08 & 0.63 & 0.59 & 0.89 & 0.40 & 0.48 & 0.55 & \textbf{0.70} & $\sim$ 5m \\
NeuralWarp & 0.49 & 0.71 & 0.38 & 0.38 & 0.79 & 0.81 & 0.82 & 1.20 & 1.06 & 0.68 & 0.66 & 0.74 & 0.41 & 0.63 & 0.51 & \textbf{0.68} & $>$ 12h \\
Neuralangelo & 0.37 & 0.72 & \cellcolor{good}0.35 & \cellcolor{nice}0.35 & 0.87 & 0.54 & \cellcolor{nice}0.53 & 1.29 & 0.97 & 0.73 & \cellcolor{nice}0.47 & 0.74 & \cellcolor{nice}0.32 & 0.41 & 0.43 & \textbf{0.61} & $\sim$ 16h \\
\midrule
SuGaR &  1.47 & 1.33 & 1.13 & 0.61 & 2.25 & 1.71 & 1.15 & 1.63 & 1.62 & 1.07 & 0.79 & 2.45 & 0.98 & 0.88 & 0.79 & \textbf{1.33} & 15--45m \\
GaussianSurfels & 0.66 & 0.93 & 0.54 & 0.41 & 1.06 & 1.14 & 0.85 & 1.29 & 1.53 & 0.79 & 0.82 & 1.58 & 0.45 & 0.66 & 0.53 & \textbf{0.88} & 6.67m \\
2DGS & 0.48 & 0.91 & 0.39 & 0.39 & 1.01 & 0.83 & 0.81 & 1.36 & 1.27 & 0.76 & 0.70 & 1.40 & 0.40 & 0.76 & 0.52 & \textbf{0.80} & 10.9m \\
GOF & 0.50 & 0.82 & \cellcolor{nice}0.37 & 0.37 & 1.12 & 0.74 & 0.73 & 1.18 & 1.29 & 0.68 & 0.77 & 0.90 & 0.42 & 0.66 & 0.49 & \textbf{0.74} & 18.4m \\
MILo & 0.43 & 0.74 & \cellcolor{best}0.34 & 0.37 & 0.80 & 0.74 & 0.70 & 1.21 & 1.22 & 0.66 & 0.62 & 0.80 & 0.37 & 0.76 & 0.48 & \textbf{0.68} & 25m \\
PGSR & \cellcolor{nice}0.36 & \cellcolor{good}0.57 & 0.38 & \cellcolor{best}0.33 & \cellcolor{nice}0.78 & \cellcolor{nice}0.58 & 0.50 & \cellcolor{nice}1.08 & \cellcolor{good}0.63 & \cellcolor{good}0.59 & \cellcolor{good}0.46 & \cellcolor{nice}0.54 & \cellcolor{best}0.30 & \cellcolor{good}0.38 & \cellcolor{best}0.34 & \cellcolor{good}\textbf{0.52} & 30m \\
GausSurf & \cellcolor{good}0.35 & \cellcolor{best}0.55 & \cellcolor{best}0.34 & \cellcolor{good}0.34 & \cellcolor{good}0.77 & \cellcolor{nice}0.58 & \cellcolor{good}0.51 & 1.10 & \cellcolor{nice}0.69 & \cellcolor{nice}0.60 & \cellcolor{best}0.43 & \cellcolor{good}0.49 & \cellcolor{nice}0.32 & \cellcolor{nice}0.40 & \cellcolor{nice}0.37 & \cellcolor{good}\textbf{0.52} & 7.2m \\
\midrule
\textbf{Ours} w/o BRDF & \cellcolor{best}0.34 & \cellcolor{good}0.57 & 0.39 & \cellcolor{good}0.34 & \cellcolor{best}0.75 & \cellcolor{best}0.51 & \cellcolor{best}0.49 & \cellcolor{best}1.03 & \cellcolor{best}0.62 & \cellcolor{best}0.57 & \cellcolor{good}0.46 & 0.56 & \cellcolor{good}0.31 & \cellcolor{good}0.38 & \cellcolor{good}0.36 & \cellcolor{best}\textbf{0.51} & 22.4m \\
\textbf{Ours} & 0.40 & \cellcolor{nice}0.59 & 0.38 & \cellcolor{nice}0.35 & \cellcolor{best}0.75 & \cellcolor{good}0.55 & 0.59 & \cellcolor{good}1.06 & \cellcolor{best}0.62 & \cellcolor{good}0.59 & \cellcolor{nice}0.47 & \cellcolor{best}0.48 & 0.34 & \cellcolor{best}0.36 & \cellcolor{nice}0.37 & \cellcolor{nice}\textbf{0.53} & 51.0m \\
\bottomrule
\end{tabular}}
\caption{Quantitative results of mesh reconstruction performance for the DTU dataset \cite{jensen2014dtu}. The Chamfer Distance (CD) $\downarrow$ for each scan of the 15 scenes in the dataset is reported. We compare our approach with SoTA neural implicit methods: VolSDF \cite{yariv2021volume}, NeuS \cite{wang2021neus}, RegSDF \cite{Zhang2022regsdf}, NeuS2 \cite{wang2023neus2}, NeuralWarp \cite{darmon2022neuralwarp}, Neuralangelo \cite{li2023neuralangelo}; and explicit methods: SuGaR \cite{guedon2023sugar}, GaussianSurfels \cite{Dai2024GaussianSurfels}, 2DGS \cite{Huang2DGS2024}, GOF \cite{Yu2024GOF}, MILo \cite{guedon2025milo}, GausSurf \cite{wang2024GausSurf}, PGSR \cite{Chen2025PGSR}. The top-three best performing results are highlighted in color for ease of visualization, with \colorbox{best}{red}, \colorbox{good}{orange}, and \colorbox{nice}{yellow} indicating 1st, 2nd, and 3rd, respectively. Note that the reconstruction runtimes are not directly comparable because these methods are trained on different GPUs.}
\label{tab-dtu-cd}
\end{table*}

\begin{equation}\label{eq-loss-ro}
    \mathcal{L}_\mathrm{ro} = \frac{1}{|\mathcal{R}|}\sum_\mathbf{p}\tanh\big(k_\mathrm{ro}(L_\mathrm{NCC}(\mathbf{p}) - \lambda_\mathrm{ref})\big)\mathcal{R}(\mathbf{p})
\end{equation}

With $\mathcal{L}_\mathrm{ro}$ regulating roughness, we apply the total variance (TV) loss $\mathcal{L}_\mathrm{tv}$ to the rendered normal map $\mathcal{N}$, where $\mathcal{R}$ acts as weights detached from the computation graph. Specifically, the roughness values are remapped to penalize varied normals at smooth regions more aggressively, and the combined depth-normal consistency term $\mathcal{L}_\mathrm{dn}$ \changed{jointly} helps refine distorted or non-watertight surfaces. Finally, we incorporate a smoothness term, $\mathcal{L}_\mathrm{sm}$, into the total loss to regulate BRDF parameters \cite{Liang2024gsir, zhu2025gsror2}. The details of these loss terms are provided in the supplementary.

\subsubsection*{Training and mesh extraction}
Our joint optimization process first trains for $5,000$ iterations to bootstrap the initialized 3D Gaussians. During this stage, all loss terms are suppressed, except $\mathcal{L}_\mathrm{rgb}$, $\mathcal{L}_\mathrm{plane}$, and the binary cross-entropy loss $\mathcal{L}_\mathrm{alpha}$ between the rendered alpha map and ground-truth masks if such masks are provided. Following the bootstrap stage, the training jointly optimizes geometric and material parameters, where all loss terms are activated and $\mathcal{L}_\mathrm{rgb}$ is replaced with $\mathcal{L}_\mathrm{pbr}$, \changed{as defined in Equation \ref{eq-L}. Note that each loss term has a corresponding weight $\lambda$, the details of which are provided in the supplementary}. Thanks to $\mathcal{L}_\mathrm{ro}$, the learning of BRDF parameters can operate in parallel with other parameters to regulate geometric details. All scenes are trained for at most $30,000$ iterations.

\begin{equation}\label{eq-L}
    \mathcal{L} = \mathcal{L}_\mathrm{plane} + \mathcal{L}_\mathrm{alpha} + \mathcal{L}_\mathrm{dn} + \mathcal{L}_\mathrm{mv} + \mathcal{L}_\mathrm{tv} + \mathcal{L}_\mathrm{sm} + \mathcal{L}_\mathrm{ro} + \mathcal{L}_\mathrm{pbr}
\end{equation}

After training, we extract triangle meshes via TSDF fusion \cite{Newcombe2011tsdf}. We render RGB-D images from all viewpoints, optionally mask depths using ground-truth alpha, and integrate them into a TSDF volume using Open3D \cite{Zhou2018Open3d}. The final mesh is obtained by marching cubes, filtered to a single connected component, and assigned per-vertex RGB colors.

\section{Experiments}
We conduct experiments to validate the effectiveness of our method using the DTU benchmark \cite{jensen2014dtu}, two scenes from the TnT dataset \cite{Knapitsch2017tnt}, and the Shiny Blender Synthetic dataset \cite{Verbin2022refnerf}. All experiments are run using a single RTX4090 GPU with 24GB VRAM, with the model split into two variants:

\begin{itemize}
    \item \textbf{Ours} w/o BRDF: training is performed without joint optimization of BRDF parameters. This helps validate the effectiveness of the new occlusion-aware filtering and multi-view normal consistency integrated into $\mathcal{L}_\mathrm{mv}$.
    \item \textbf{Ours}: the joint optimization process as described in Section \ref{sec-med-gs2m}.
\end{itemize}

Table \ref{tab-dtu-cd} compares the performance of our model with SoTA implicit and explicit mesh reconstruction methods using the DTU benchmark. As with previous works, we report the Chamfer distance (CD) for each of the 15 scenes in the dataset, together with the average runtime. It is clear that all explicit methods, including GS-2M, consume significantly less training time and computing resources compared to implicit methods. Our joint training pipeline, however, takes roughly double the reconstruction time of the variant without BRDF optimization. This is due to the deferred rendering step and an extra overhead resulting from the computation of our under-optimized $\mathcal{L}_\mathrm{ro}$. Nevertheless, the quantitative results show that both variants perform on par with SoTA explicit surface reconstruction methods and outperform all neural implicit methods, maintaining the reconstruction quality even when jointly optimizing with BRDF parameters and the PBR pipeline. In some specific scenes, we even outperform the current SoTA explicit methods, proving the effectiveness of the modified multi-view constraints.

\begin{figure*}
    \centering
    \includegraphics[width=1\linewidth]{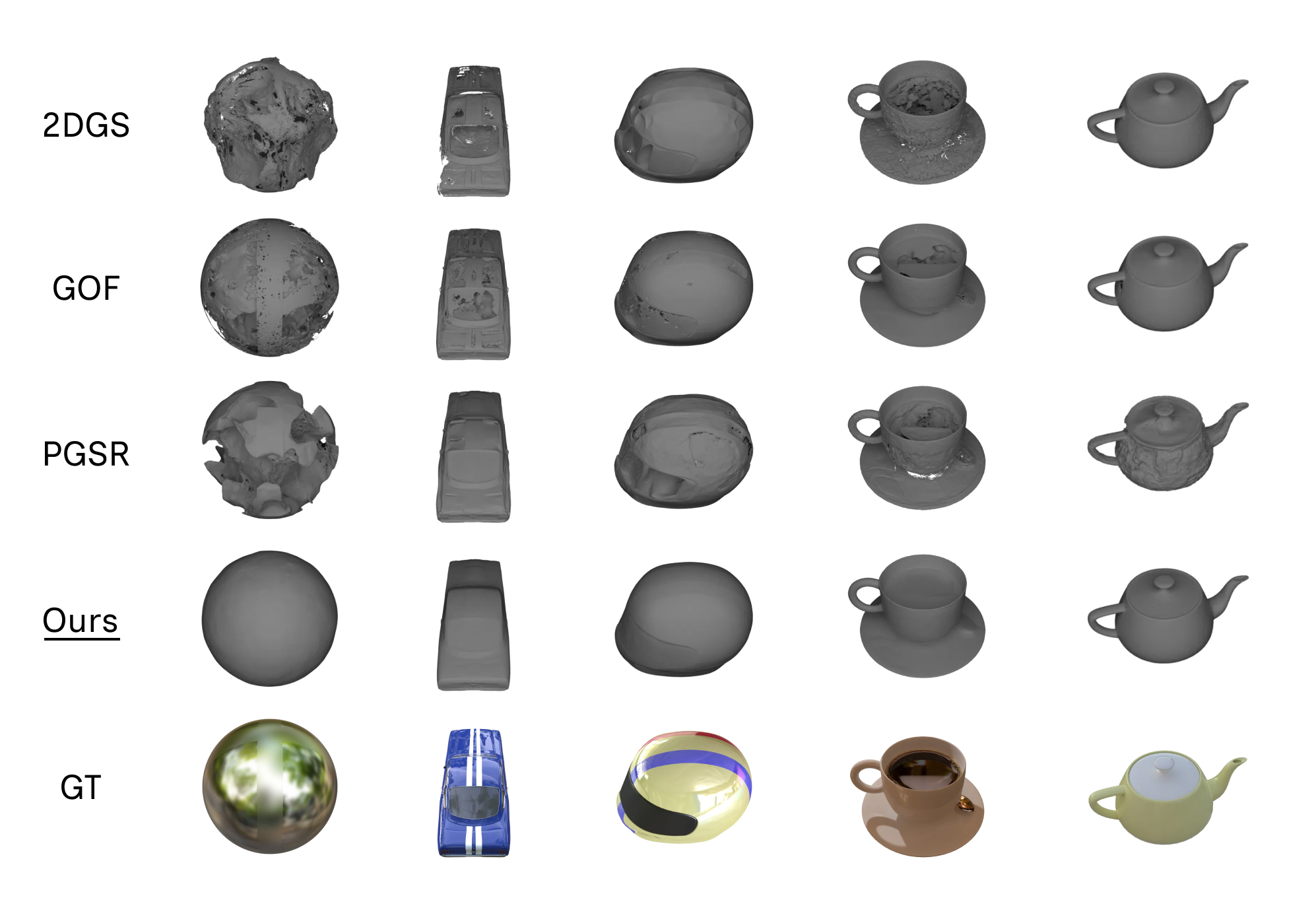}
    \caption{Qualitative comparisons for the Shiny Blender Synthetic dataset \cite{Verbin2022refnerf}. We compare the reconstructed meshes of our full model with current SoTA neural explicit methods: 2DGS \cite{Huang2DGS2024}, GOF \cite{Yu2024GOF}, PGSR \cite{Chen2025PGSR}. Due to the limited appearance modeling, these methods sacrifice geometric details for view-dependent effects caused by specular surfaces, resulting in non-watertight or distorted meshes. In contrast, our method produces more uniform surfaces thanks to the joint constraints of the $\mathcal{L}_\mathrm{ro}$, $\mathcal{L}_\mathrm{tv}$, and $\mathcal{L}_\mathrm{dn}$ loss terms. Moreover, GS-2M still maintains the reconstruction quality on the DTU benchmark, making it a unified solution for both mesh reconstruction and material decomposition. Please refer to the supplementary material and our project page for more experimental results.}
    \label{fig-shiny-meshes}
\end{figure*}

\changed{Table \ref{tab-dtu-cd} also reveals that our joint optimization framework with BRDF parameters introduces an inevitable cost–benefit trade-off. While the deferred PBR pipeline and roughness regularization roughly double the training time, they do not always improve the Chamfer distance; in several scenes, our model without BRDF marginally outperforms the full model. We attribute this to shape–reflectance ambiguities and the additional degrees of freedom introduced by BRDF parameters, which can slightly hinder geometric convergence in scenes with weak or inconsistent reflective cues. Nevertheless, these differences are small and scene-dependent, and the joint model remains competitive with SoTA explicit methods while additionally recovering physically meaningful material properties, making it more suitable for photorealistic rendering and downstream PBR tasks.}

\begin{table*}[t]
\centering
\resizebox{\textwidth}{!}{
\begin{tabular}{lcccccccccccccccc}
\toprule
\textbf{PSNR} $\uparrow$ & 24 & 37 & 40 & 55 & 63 & 65 & 69 & 83 & 97 & 105 & 106 & 110 & 114 & 118 & 122 & Mean \\
\midrule
RegSDF & 24.78 & 23.06 & 23.47 & 22.21 & 28.57 & 25.53 & 21.81 & 28.89 & 26.81 & 27.91 & 24.71 & 25.13 & 26.84 & 21.67 & 28.25 & \textbf{25.31} \\
NeuS & 26.49 & 26.17 & 27.66 & 27.78 & 30.63 & 27.42 & 25.38 & 30.00 & 26.40 & 29.63 & 25.87 & 28.82 & 28.80 & 27.36 & 31.19 & \textbf{28.00} \\
NeuS2 & 28.44 & 27.14 & 29.70 & 29.67 & 31.75 & 27.83 & 24.84 & 31.24 & 26.86 & 30.57 & 26.05 & 28.93 & 28.98 & 27.82 & 32.48 & \textbf{28.82} \\
VolSDF & 26.28 & 25.61 & 26.55 & 26.76 & 31.57 & 31.50 & 29.38 & 33.23 & 28.03 & 32.13 & 33.16 & 31.49 & 30.33 & 34.90 & 34.75 & \textbf{30.38} \\
Neuralangelo & 30.64 & 27.78 & \cellcolor{best}32.70 & \cellcolor{best}34.18 & \cellcolor{good}35.15 & \cellcolor{best}35.89 & 31.47 & \cellcolor{good}36.82 & 30.13 & \cellcolor{best}35.92 & \cellcolor{good}36.61 & 32.60 & 31.20 & \cellcolor{best}38.41 & \cellcolor{best}38.05 & \cellcolor{nice}\textbf{33.84} \\
% \midrule
PGSR & \cellcolor{good}32.20 & \cellcolor{nice}28.07 & 31.10 & \cellcolor{nice}33.41 & 33.99 & 33.22 & \cellcolor{good}32.16 & 32.83 & \cellcolor{nice}31.31 & 33.82 & \cellcolor{nice}36.03 & \cellcolor{nice}34.64 & \cellcolor{good}32.71 & 37.33 & 37.09 & \textbf{33.33} \\
% 2DGS & \cellcolor{best}34.16 & \cellcolor{best}31.45 & \cellcolor{best}33.78 & \cellcolor{nice}33.65 & \cellcolor{best}36.99 & \cellcolor{good}34.48 & \cellcolor{best}33.05 & 32.66 & 24.22 & \cellcolor{good}34.23 & 36.19 & 34.94 & \cellcolor{best}32.90 & 37.37 & 37.64 & \cellcolor{good}\textbf{33.85} \\
% GOF & 34.45 & 32.16 & 34.44 & 35.58 & 37.73 & 33.94 & 33.62 & 31.43 & 31.00 & 35.95 & 38.60 & 35.51 & 34.42 & 39.18 & 37.94 & \textbf{35.06} \\
% GaussianSurfels & 33.17 & 30.25 & 33.34 & 35.11 & 36.99 & 34.36 & 32.56 & 39.07 & 31.89 & 35.69 & 37.52 & 35.57 & 33.67 & 39.84 & 40.21 & \textbf{35.28} \\
\midrule
% \textbf{Ours} w/o BRDF & 33.70 & 30.00 & 33.08 & 34.17 & 35.56 & 33.93 & 32.52 & 33.87 & 31.65 & 34.90 & 36.99 & 35.56 & 33.19 & 38.37 & 38.27 & \textbf{34.38} \\
\textbf{Ours} w/o BRDF & \cellcolor{best}33.16 & \cellcolor{best}30.47 & \cellcolor{nice}31.54 & \cellcolor{good}34.04 & \cellcolor{best}35.78 & \cellcolor{good}34.41 & \cellcolor{best}32.47 & \cellcolor{nice}33.76 & \cellcolor{good}31.58 & \cellcolor{nice}34.55 & \cellcolor{best}36.74 & \cellcolor{good}35.54 & \cellcolor{best}33.20 & \cellcolor{good}38.30 & \cellcolor{nice}37.73 & \cellcolor{best}\textbf{34.22} \\
\textbf{Ours} & \cellcolor{nice}31.62 & \cellcolor{good}29.30 & \cellcolor{good}31.55 & \cellcolor{nice}33.41 & \cellcolor{nice}35.05 & \cellcolor{nice}33.32 & \cellcolor{nice}31.77 & \cellcolor{best}37.44 & \cellcolor{best}31.87 & \cellcolor{good}34.62 & 34.48 & \cellcolor{best}35.73 & \cellcolor{nice}32.20 & \cellcolor{nice}37.70 & \cellcolor{good}37.85 & \cellcolor{good}\textbf{33.86} \\
% \textbf{Ours} \\
\bottomrule
\end{tabular}}
\caption{Quantitative results of novel-view synthesis (NVS) performance for the DTU dataset \cite{jensen2014dtu}. We report the Peak Signal-to-Noise Ratio (PSNR) $\uparrow$ for each scene of the 15 scenes in the dataset. We compare our approach with SoTA surface reconstruction methods: VolSDF \cite{yariv2021volume}, NeuS \cite{wang2021neus}, RegSDF \cite{Zhang2022regsdf}, NeuS2 \cite{wang2023neus2}, Neuralangelo \cite{li2023neuralangelo}, and PGSR \cite{Chen2025PGSR}. Except for neural implicit methods, all NVS metrics are reproduced using the published code and evaluated on unmasked images. The top-three best performing results are highlighted, with \colorbox{best}{red}, \colorbox{good}{orange}, and \colorbox{nice}{yellow} indicating 1st, 2nd, and 3rd, respectively. Incorporating multi-view normal consistency and occlusion-aware filtering enhances NVS quality while preserving the fidelity of the extracted mesh. Our method outperforms all SoTA reconstruction approaches, thanks to these additions. Note that the NVS performance for the DTU dataset is evaluated on the whole training data, aligning with previous works.}
\label{tab-exp-dtu-psnr}
\end{table*}

The major advantage of our framework, however, is the ability to reconstruct reflective objects, as shown in Figure \ref{fig-shiny-meshes}. We compare the reconstructed mesh of our joint optimization solution with meshes produced by recent SoTA neural explicit methods. These qualitative results are produced using code provided by the authors of the respective works. For a fair comparison, we match the number of training iterations and modify the published code to handle white backgrounds without generating floaters. The qualitative comparisons validate the effectiveness of our joint optimization framework. Specifically, the multi-view self-supervision term $\mathcal{L}_\mathrm{ro}$ helps detect diffuse and specular regions, driving the TV loss $\mathcal{L}_\mathrm{tv}$ to force smooth normals accordingly via roughness weighting. These smooth normals, in turn, regulate the geometric details thanks to the depth-normal consistency loss $\mathcal{L}_\mathrm{dn}$. These terms cooperate in a unified optimization framework to deliver high-quality reconstruction results for both diffuse and specular surfaces.

As with previous neural implicit methods, we report the novel-view synthesis (NVS) performance of our model using the DTU benchmark, measured by the peak signal-to-noise ratio (PSNR) between the rendered and ground-truth images. For a fair comparison, we evaluate the NVS metrics on the training set, aligning with previous works. Table \ref{tab-exp-dtu-psnr} compares the NVS quality of our method with neural implicit works that reported the same metric. Additionally, we reproduce the NVS performance of PGSR \cite{Chen2025PGSR} using their publicly available code. The table reveals that enhancing the multi-view loss term with normal consistency and filtering occluded samples helps our solution outperform all SoTA surface reconstruction methods in terms of NVS quality. For the PBR variant of our model, there is a slight drop in NVS performance, which we attribute to the limited constraints for the environment lighting and albedo. Specifically, the optimization process of our method still leaves abundant freedom for these two components, resulting in noise artifacts in the material decomposition results. Nevertheless, our joint optimization with BRDF parameters still keeps the NVS quality on par with current SoTA methods.

Finally, we extend our experiments to unbounded scenes using the TnT dataset \cite{Knapitsch2017tnt}. Unlike DTU and Shiny Blender Synthetic, the TnT dataset features sequences acquired outside controlled labs under realistic conditions, including both challenging outdoor scenes and complex indoor environments. Since our joint framework adopts the PBR pipeline designed for object-centric scenes, we only use the TnT dataset to validate the w/o BRDF variant. \changed{This design choice reflects our intentional focus on object-centric reconstruction, where controlled lighting, limited background clutter, and stable material properties make physically based shading both meaningful and reliable. Nevertheless, even outside this intended regime, our framework retains practical advantages, including fast convergence, low memory footprint relative to neural implicit methods, and strong geometric consistency from the modified multi-view constraints}. As with previous works, we report the F1 scores, balancing the precision and recall tests between the reconstructed mesh and the ground-truth scan. Table \ref{tab-exp-tnt-f1} compares the reconstruction performance of our model with SoTA neural implicit and explicit methods for the Barn and Truck scenes. The comparison reveals that the new modification to the multi-view geometric term maintains the quality of the reconstructed mesh even without the exposure compensation module that PGSR employed to benchmark with the TnT dataset. However, we restrict our experiments to these two scenes, as the remaining ones cause out-of-memory errors due to the lack of control over generating new Gaussians, especially for scenes with overwhelming background details. We discuss the limitations of our approach and promising solutions in the next section.

\begin{table}[t]
\centering
\resizebox{\linewidth}{!}{
\begin{tabular}{l|ccc|cccc|c}
\toprule
\textbf{F1-Score} $\uparrow$ & NeuS & Geo-NeuS & Neuralangelo & SuGaR & 2DGS & GOF & PGSR & \textbf{Ours} w/o BRDF \\
\midrule
Barn        & 0.29 & 0.33 & \cellcolor{best}0.70 & 0.14 & 0.36 & 0.51 & \cellcolor{good}0.66 & \cellcolor{nice}0.57  \\
% Caterpillar & 0.29 & 0.26 & 0.36 & 0.16 & 0.23 & 0.41 & 0.41 \\
% Courthouse  & 0.17 & 0.12 & 0.28 & 0.08 & 0.13 & 0.28 & 0.21 \\
% Ignatius    & 0.83 & 0.72 & 0.89 & 0.33 & 0.44 & 0.68 & 0.80 \\
% MeetingRoom & 0.24 & 0.20 & 0.32 & 0.15 & 0.16 & 0.28 & 0.29 \\
Truck       & 0.45 & 0.45 & 0.48 & 0.26 & 0.26 & \cellcolor{nice}0.58 & \cellcolor{good}0.66 & \cellcolor{best}0.67 \\
\midrule
Mean        & 0.37 & 0.39 & \cellcolor{nice}0.59 & 0.15 & 0.31 & 0.55 & \cellcolor{best}0.66 & \cellcolor{good}0.62 \\
% Time        & $>$24h & $>$24h & $>$24h & $>$2h & 34.2m & $>$2h & 1.2h \\
\bottomrule
\end{tabular}}
\caption{Quantitative results of mesh reconstruction performance for the Truck and Barn scenes from the TnT dataset \cite{Knapitsch2017tnt}, where the F1-score $\uparrow$ is reported. We compare our reconstructed meshes with those of NeuS \cite{wang2021neus}, Geo-NeuS \cite{Fu2022GeoNeus}, Neuralangelo \cite{li2023neuralangelo}, SuGaR \cite{guedon2023sugar}, 2DGS \cite{Huang2DGS2024}, GOF \cite{Yu2024GOF}, and PGSR \cite{Chen2025PGSR}. The top-three best performing results are highlighted, with \colorbox{best}{red}, \colorbox{good}{orange}, and \colorbox{nice}{yellow} indicating 1st, 2nd, and 3rd, respectively.}
\label{tab-exp-tnt-f1}
\end{table}

\section{Ablation, Limitations, Future Work}
\begin{figure}
    \centering
    \includegraphics[width=1\linewidth]{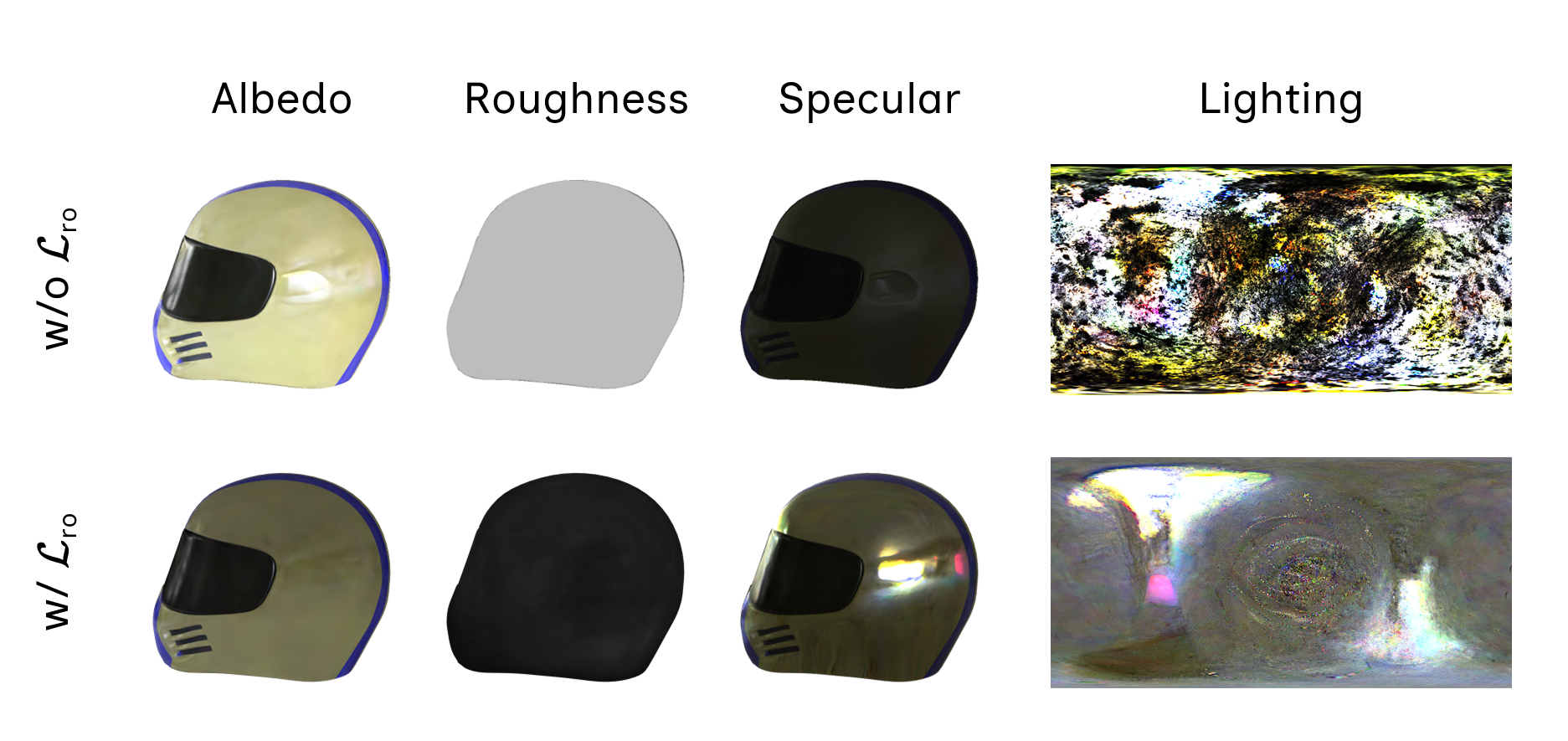}
    \caption{Training without $\mathcal{L}_\mathrm{ro}$ regulating roughness (top row) results in noisy lighting and albedo bleeding. In comparison, the use of $\mathcal{L}_\mathrm{ro}$ (bottom row) enables the optimization process to learn the correct roughness values at shiny regions.}
    \label{fig-ablation-roughness}
\end{figure}

For ablation, we study the effectiveness of the enhanced multi-view geometric term and the proposed roughness supervision loss. Since our framework is largely constructed from components introduced in previous works \cite{Chen2025PGSR, Huang2024visbility, Liang2024gsir, zhu2025gsror2}, please refer to their respective papers for deeper insights. Figure \eqref{fig-ablation-roughness} compares the appearance decomposition results when training with and without the multi-view roughness supervision term ($\mathcal{L}_\mathrm{ro}$). The proposed loss reduces noise artifacts in the captured lighting and prevents specular highlights from bleeding into the albedo. Conversely, relying solely on $\mathcal{L}_\mathrm{pbr}$ is extremely under-constrained, contaminating learned BRDF parameters and environment lighting.

\begin{table}[h]
    \centering
    \begin{tabular}{l|c|c}
         DTU benchmark & CD $\downarrow$ & PSNR $\uparrow$  \\
         \midrule
         \textbf{Ours} w/o mv normal & 0.58 & 26.73 \\
         \textbf{Ours} w/o filtering & 0.53 & 33.76 \\
         \textbf{Ours} & 0.53 & 33.86 \\
    \end{tabular}
    \caption{Ablation study for the modified multi-view geometric constraint, conducted on the DTU benchmark \cite{jensen2014dtu}. Introducing multi-view normal consistency substantially increases reconstruction and NVS performance, while occlusion filtering further enhances NVS quality.}
    \label{tab-ablation-mv-geo}
\end{table}

Table \ref{tab-ablation-mv-geo} studies the effect of two modifications made to the multi-view geometric constraint $L_\mathrm{g}$, using the DTU benchmark. The ablation reveals that the addition of multi-view normal consistency to the $L_\mathrm{g}$ term substantially enhances both surface reconstruction and NVS performance. As demonstrated in Figure \ref{fig-mv-pgsr}, the new $L_\mathrm{g}$ also helps achieve more consistent geometry in regions with high-frequency textures. The integration of occlusion-aware filtering, on the other hand, only causes a slight improvement in NVS quality. Nevertheless, we recognize certain limitations of our method that provide opportunities for future enhancement.

% From left to right: the rendered albedo, roughness, specular, and captured lighting.

% Our framework successfully maintains surface reconstruction performance for diffuse objects and simultaneously handles highly reflective ones.

\paragraph*{Under-constrained albedo and lighting}
Despite the effectiveness of the roughness supervision term, the optimization process of our method still lacks constraints to supervise albedo and lighting. Although contemporary works handle this ill-posed problem with encoders--decoders \cite{Zhang2025refgs} or pre-trained priors \cite{glossygs}, we prefer self-supervision strategies similar to $\mathcal{L}_\mathrm{ro}$. In other words, finding reliable features to guide albedo and lighting without heavily relying on neural components is an interesting direction that we regard as a potential improvement for decomposing appearance.

\begin{figure}
    \centering
    \includegraphics[width=1\linewidth]{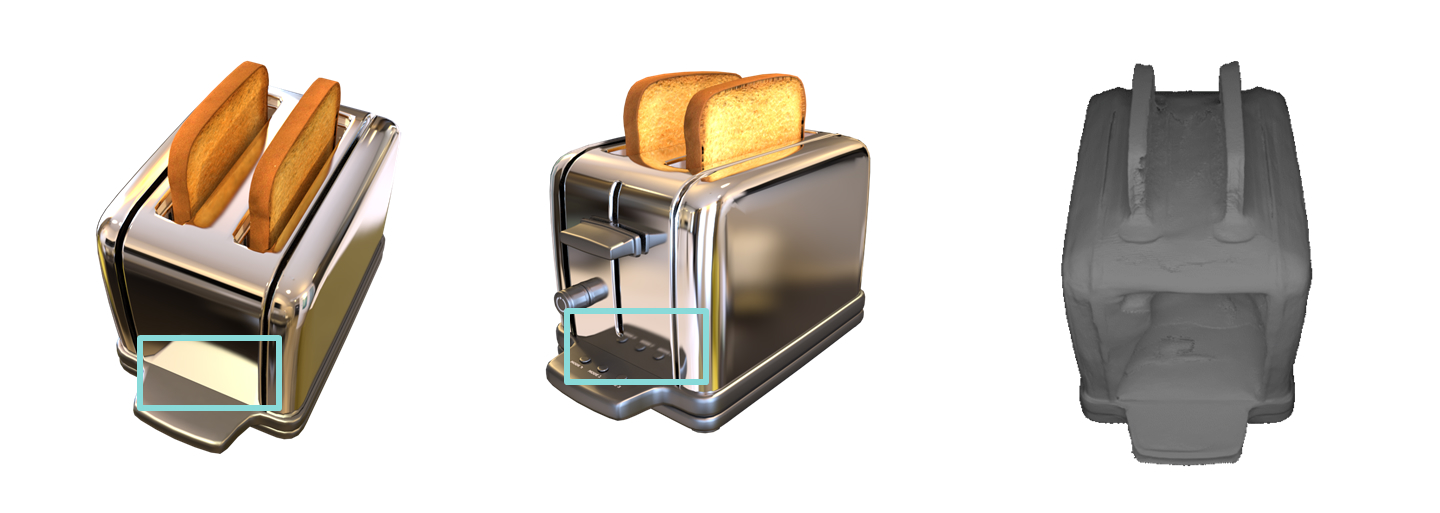}
    \caption{The reconstructed mesh of the Toaster scene from the Shiny  Blender Synthetic dataset \cite{Zhang2025refgs, 3dgrt2024}. We fail to recover this scene due to the limited capability of the employed shading model.}
    \label{fig-limit-self}
\end{figure}

\paragraph*{Self-reflection and self-shadowing}
The incorporation of BRDF parameters helps cope with reflective surfaces, yet our implementation cannot faithfully decompose the appearances of objects exhibiting self-reflection. Figure \ref{fig-limit-self} illustrates our method's limitation in reconstructing such objects, where self-reflecting regions pose challenges to the optimization process. These appearance properties require ray-based approaches \cite{Gao2023relightable, 3dgrt2024} to model indirect lighting, and we regard this as a promising enhancement to our joint supervision framework.

\begin{figure}[h]
    \centering
    \includegraphics[width=1\linewidth]{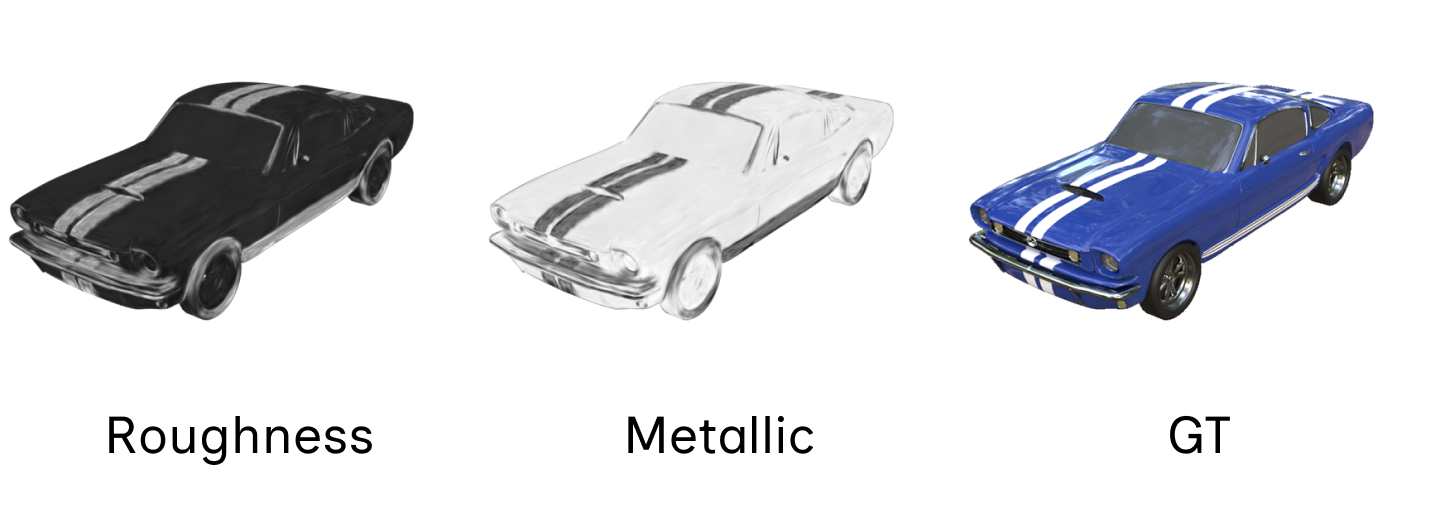}
    \caption{The approximation of metallic from roughness is not coherent with the appearance properties of some objects. Notice how the white stripes of the car in the rendered metallic map are approximated as dielectric, where they should be metal.}
    \label{fig-limit-metallic}
\end{figure}

\paragraph*{Learnable metallic}
Currently, we approximate the metallic component of the BRDF as $m = 1 - \rho$, which is not ideal in some scenes. Figure \ref{fig-limit-metallic} illustrates one such case, where metallic values along the white stripes are not coherent with the car surface because they are derived from the roughness map. To this end, making the metallic parameter learnable is a preferred improvement, yet this would require advanced supervision strategies to avoid overfitting, which we leave as a future perspective.

\paragraph*{Better densification}
As discussed in the previous section, we encounter out-of-memory errors when reconstructing unbounded scenes \cite{Knapitsch2017tnt}. This is the direct consequence of using the unmodified adaptive density control (ADC) of 3DGS \cite{kerbl3dgs}, where excessive Gaussians are generated due to the vast background details. Moreover, the shading function we employ for our method is designed for object-centric scenes, making the joint optimization pipeline unfavorable for these scenarios. We are interested in adopting recent outstanding works \cite{Kheradmand2024gsmcmc} to help with reconstructing large-scale scenes.

\paragraph*{Object-centric scenes}
\changed{Our framework is primarily designed for object-centric reconstruction because the adopted PBR pipeline and deferred shading formulation are not well suited for large-scale or unbounded scenes with complex illumination and extensive background geometry. While our method can still recover reasonable geometry in such cases, the lack of explicit modeling for global illumination and environment lighting limits both stability and scalability. We therefore position our approach as a high-fidelity object-level reconstruction method and regard extensions toward unbounded scenes as an important direction for future work.}

\section{Conclusion}
We have presented and discussed our \changed{material-aware, joint-optimization} framework for high-fidelity mesh reconstruction \changed{3D Gaussian splatting}. We started introducing the emergence of neural rendering as a faithful paradigm to replace manual, laborious work in visual computing domains. While neural implicit methods promise their effectiveness in addressing a large number of vision tasks, we chose to pursue Gaussian splatting approaches due to their explicit and resource-friendly nature. On assessing the capabilities of SoTA neural explicit methods for mesh reconstruction, we discovered that they struggle to reconstruct highly reflective surfaces due to the lack of appearance modeling. We therefore propose a joint optimization solution with BRDF parameters independent of external factors and a novel roughness supervision strategy based solely on multi-view photometric variation. The experiments show that our framework maintains the reconstruction performance with SoTA for diffuse objects while simultaneously handling highly reflective ones, making it suitable for both mesh reconstruction and material decomposition. Despite still carrying several limitations \changed{and ideal for object-centric scenes}, we believe our solution is a starting point in unifying both tasks, aiming toward high-fidelity surface and appearance reconstruction.

%-------------------------------------------------------------------------

% acknowledgement
\section*{Acknowledgements}
This research was carried out within the COSI Erasmus Mundus Master’s Program. We acknowledge the support of Carl Zeiss AG for providing computational resources and research infrastructure.

% bibtex
\bibliographystyle{eg-alpha-doi} 
\bibliography{egbibsample}       

% biblatex with biber
% \printbibliography                

\includepdf[pages=-]{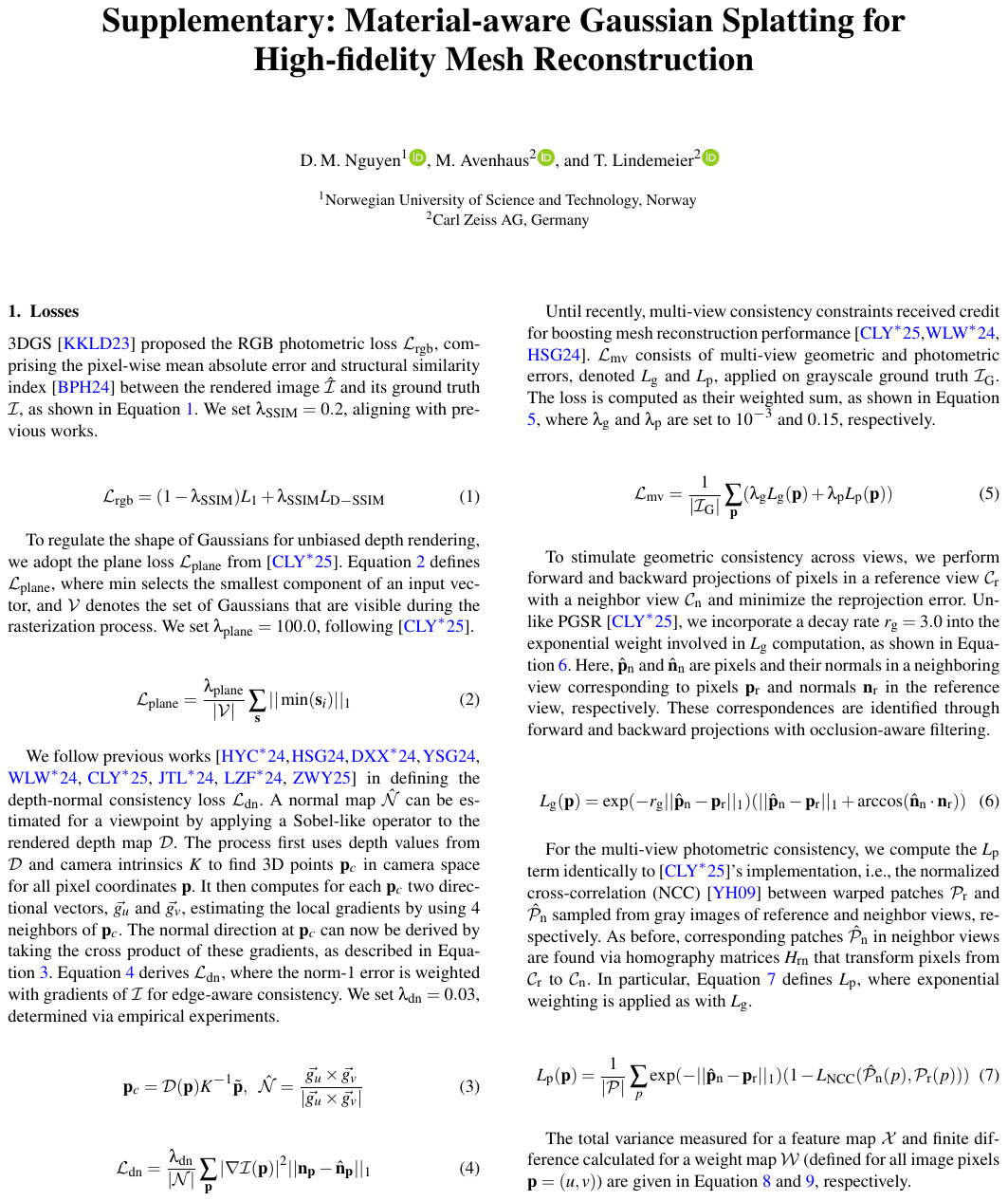}

\end{document}